\documentclass[lettersize,journal]{IEEEtran}
\usepackage{amsmath,amsfonts}
\usepackage{algorithmic}
\usepackage{algorithm}
\usepackage{array}
\usepackage{textcomp}
\usepackage{stfloats}
\usepackage{url}
\usepackage{verbatim}
\usepackage{graphicx}
\usepackage{cite}

\usepackage{multirow}
\usepackage{bbding}
\usepackage{subfigure}
\usepackage{booktabs}
\usepackage{color} 
\usepackage{siunitx}
\usepackage{colortbl}
\definecolor{mygray}{gray}{.9}
\definecolor{mypurple}{RGB}{128,0,128}
\begin{document}

\title{\color{black}{Multi-Dimensional Quality Assessment for Text-to-3D Assets: Dataset and Model}}



\author{
Kang Fu*, Huiyu Duan*, Zicheng Zhang, Xiaohong Liu, \emph{Member, IEEE,}\\  Xiongkuo Min$^{\dagger}$, \emph{Member, IEEE,} Jia Wang, and Guangtao Zhai$^{\dagger}$, \emph{Fellow, IEEE
} 


\IEEEcompsocitemizethanks{\IEEEcompsocthanksitem Kang Fu, Huiyu Duan, Zicheng Zhang, Xiaohong Liu, Xiongkuo Min, Jia Wang and Guangtao Zhai are with Shanghai Jiao Tong University, 200240 Shanghai, China. E-mail:\{fuk20\-20, huiyuduan, zzc1998, xiaohongliu, minxiongkuo, jiawang, zhaiguangtao\}@sjtu.edu.cn. 
This work was supported in part by the National Key R\&D Program of China under Grant 2021YFE0206700, in part by the National Natural Science Foundation of China under Grants 62401365, 62271312, 62225112, 62132006, and in part by the Shanghai Pujiang Program under Grant 22PJ1407400. * Equal Contributions. $\dagger$ Corresponding Authors.\protect}
}



\maketitle

\begin{abstract}

Recent advancements in text-to-image (T2I) generation have spurred the development of text-to-3D asset (T23DA) generation, leveraging pretrained 2D text-to-image diffusion models for text-to-3D asset synthesis. Despite the growing popularity of text-to-3D asset generation, its evaluation has not been well considered and studied. However, given the significant quality discrepancies among various text-to-3D assets, there is a pressing need for quality assessment models aligned with human subjective judgments. To tackle this challenge, we conduct a comprehensive study to explore the T23DA quality assessment (T23DAQA) problem in this work from both subjective and objective perspectives. Given the absence of corresponding databases, we first establish the largest text-to-3D asset quality assessment database to date, termed the AIGC-T23DAQA database. This database encompasses 969 validated 3D assets generated from 170 prompts via 6 popular text-to-3D asset generation models, and corresponding subjective quality ratings for these assets from the perspectives of quality, authenticity, and text-asset correspondence, respectively. Subsequently, we establish a comprehensive benchmark based on the AIGC-T23DAQA database, and devise an effective T23DAQA model to evaluate the generated 3D assets from the aforementioned three perspectives, respectively. Specifically, the proposed method utilizes the projection videos of text-to-3D assets to extract 3D shape, texture and text-asset correspondence features, then fuses them to calculate the final three preference scores respectively. Extensive experimental results demonstrate the effectiveness of the proposed T23DAQA method in evaluating the quality of AI generated 3D asset, which is more consistent with human perception. To the best of our knowledge, this is the first work that studies the problem of text-guided 3D generation quality assessment, and  The database is released at \url{https://github.com/ZedFu/T23DAQA}.

\end{abstract}

\begin{IEEEkeywords}
text-to-3D asset generation, subjective quality assessment, objective quality assessment, artificial intelligence generated content (AIGC)
\end{IEEEkeywords}

\section{Introduction}

\begin{figure}[t]
    \centering
    \includegraphics[width = 0.48\textwidth]{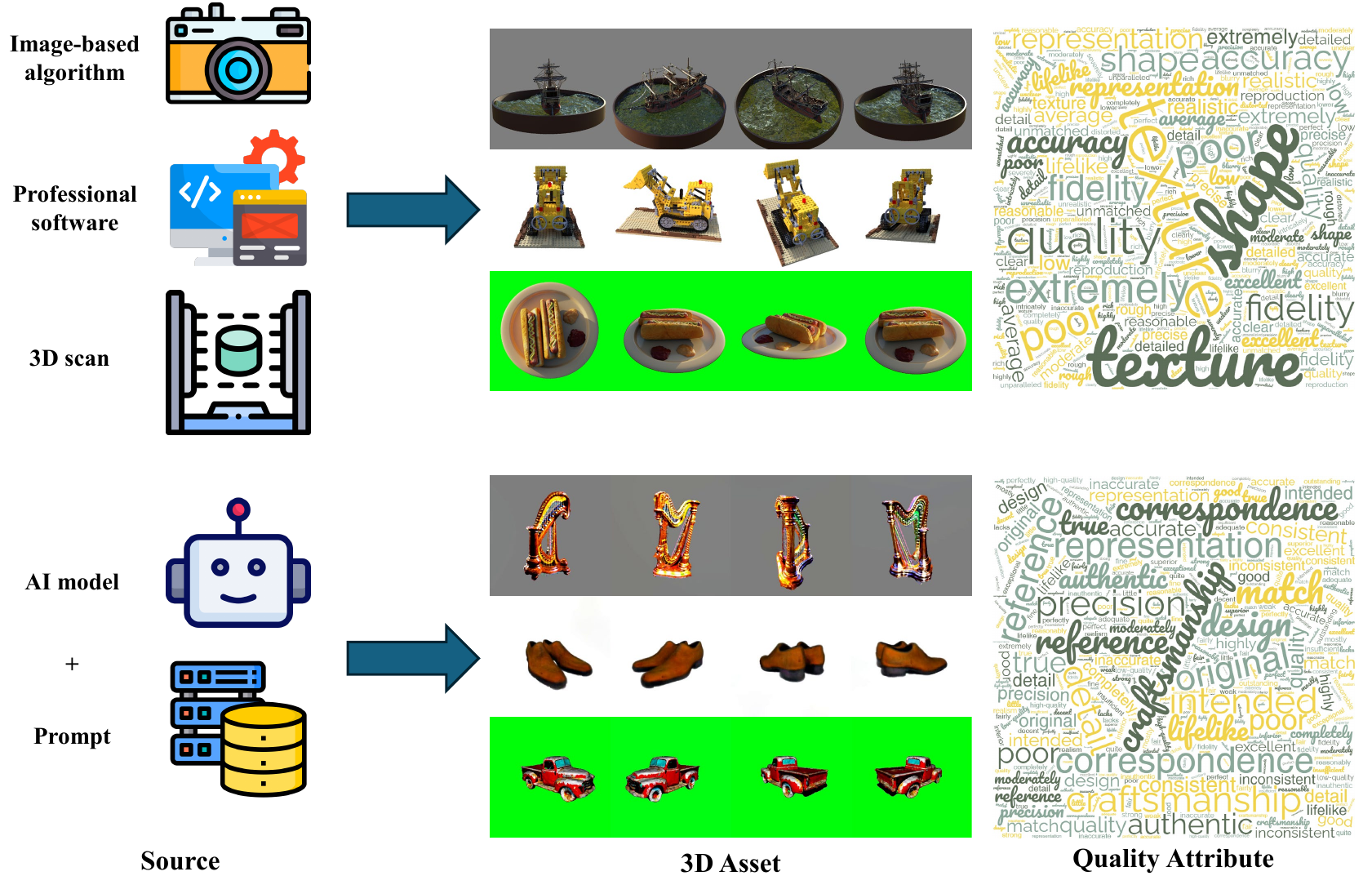}
    \caption{Illustration of the difference between traditional 3d asset and AI generated 3d asset, whose perceptual quality are affected by different attributes.}
    \label{fig:thumbnail}
    \vspace{-0.5cm}
\end{figure}

\IEEEPARstart{T}{he} 3D asset generation has long been an important task in the field of computer vision (CV) and artificial intelligence (AI), which pursues high-quality and automatic 3D model or view synthesis \cite{henzler2019escaping, tsalicoglou2023textmesh}. The recent advances in text-to-image generation via diffusion models have spurred the development of numerous text-to-3D asset generation methodologies, exemplified by works including Dreamfusion \cite{poole2022dreamfusion}, Prolificdreamer \cite{wang2024prolificdreamer}, \textit{etc}. However, the text-to-3D asset synthesis is influenced by various factors such as prompts, and techniques, leading to diverse perceptual qualities that directly impact user experience. Consequently, there is a crucial need for a subjective-consistent quality assessment framework to evaluate text-to-3D assets. However, existing quality assessment models fail to adequately address this task. As shown in Fig. \ref{fig:thumbnail}, on one hand, distortions introduced by text-to-3D asset generation models, including unreal structures, unreasonable components, discontinuous views, are significantly different from those encountered in traditional 3D asset, which invalidates traditional quality assessment methods. On the other hand, conventional quality assessment models do not take the alignment between text and 3D asset into consideration, which is a pivotal evaluation aspect for text-to-3D assets.

Current text-to-3D asset generation models generally uses image fidelity evaluation metrics such as Inception Score (IS) \cite{gulrajani2017improved} and Fréchet Inception Distance (FID) \cite{heusel2017gans} to assess the quality of text-to-3D assets. However, these metrics cannot evaluate the fidelity, quality and text-image correspondence of a single generated image. Moreover, previous quality assessment metrics designed for natural images, omnidirectional images, natural videos, user generated videos, point clouds, meshes \textit{etc.}\cite{zhou2024blind,chen2024dynamic,sun2022a,wu2023dover}, may not generalize well for assessing text-to-3D assets. There are two main reasons for this: 1) Previous quality assessment methods can only predict the quality aspect of the generated asset, while ignoring the authenticity aspect and the association between the prompt and the generated 3D asset; 2) The distortions existed in text-to-3D asset, such as floating artifacts and multiple similar planes, are different from the common distortions existed in traditional 3D models, such as noise and compression, which makes most traditional quality assessment methods unable to generalize well to assess the quality of T23DA. Several text-to-3D generation works also conduct the user studies, which let volunteers choose the better generated 3D assets, to validate the effectiveness of a generation framework. However, user studies are time consuming and inefficient, which further strengths the importance of developing an objective perception evaluation algorithm for text-to-3D assets. In the same time, T23DAQA has many potential applications in real-world scenarios: 1) it has the potential to be used to optimize the perceptual quality of generated 3D asset as a loss function in the training of a text-to-3D asset model. 2) Nowadays, many T23DA companies have emerged to assist game designers and filmmakers in creating 3D asset, such as Genie \cite{genie} and Meshy\cite{meshy}. However, the generated 3D asset is not always excellent and usually requires to select the best 3D asset from multiple generated results. T23DAQA can automatically filter out generated 3D asset with better perceptual quality.
\begin{figure*}[t]
    \centering
    \includegraphics[width = 0.98\textwidth]{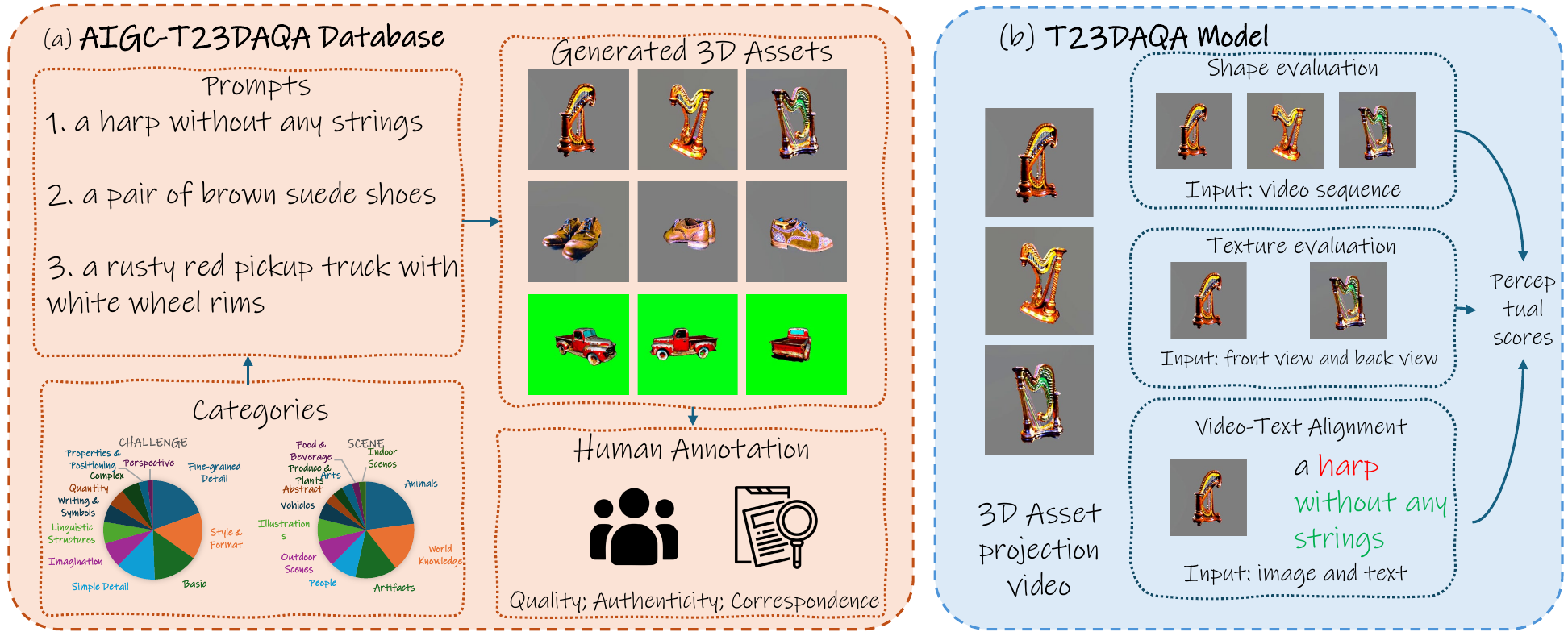}
    \caption{\color{black}{An Overview of the established AIGC-T23DAQA database and the proposed T23DAQA method. AIGC-T23DAQA database is the first and the largest text-to-3d assets quality assessment database. This database encompasses 969 validated 3D assets generated from 170 prompts via 6 popular text-to-3D asset generation models, and corresponding subjective quality ratings. In addition, we popose a T23DAQA method to predict the text-to-3D asset quality from three aspects: shape, texture, and correspondence. The proposed method achieves the state-of-the-art performance in evaluating the perceptual attributes of text-to-3d assets.}}
    \label{fig:spotlight}
    \vspace{-0.5cm}
\end{figure*}
As shown in Fig. \ref{fig:spotlight}, in order comprehensively and accurately evaluate text-to-3D assets, we conduct both subjective and objective assessment studies in this work. Firstly, we establish the largest-scale subjective text-to-3D asset quality assessment dataset to date, named AIGC-T23DAQA database. This dataset comprises 969 text-to-3D assets generated by six distinct text-to-3D methods using 170 text prompts. Based on the generated 3D assets, a subjective experiment is conducted to collect the quality, authenticity, and text-asset correspondence ratings, respectively, which are further processed to obtain mean opinion scores (MOSs).  To the best of our knowledge, this is the first database for text-to-3D generated asset evaluation from multiple perspectives.
Based on the established AIGC-T23DAQA database, we propose a novel model equipped with multi-modality foundation models for better text-to-3D asset quality assessment, which is named as T23DAQA model. Since the most of recent text-to-3D asset generation methods employ neural radiation fields (NeRF) to represent 3D asset, which are stored in multilayer perceptrons (MLPs) or voxels. The NeRF-based representation generally lacks explicit 3D models, which poses challenges for 3D quality assessment. So our proposed method is a projection-based quality evaluation algorithm that extracts perceptual quality features from three aspects, including: shape, texture, and text-asset correspondence, and then fuse the extracted features to predict quality, authenticity, and text-asset correspondence scores. Based on the constructed AIGC-T23DAQA database, we establish a benchmark for it including many SOTA quality assessment methods and validate the effectiveness of the proposed method on this benchmark. Experimental results demonstrate that our proposed method achieves the best performance compared to these state-of-the-art methods for evaluating text-to-3D assets, which manifests the superiority of the proposed model.

\textcolor{black}{In summary, the motivation of conducting this work is that there are many text-based 3D asset generation methods and corresponding generated assets, however, the existing quality assessment algorithms cannot well evaluate the performance of these models and the quality of the generated assets. As the first text-to-3D asset quality assessment work, the proposed database can be used to develop corresponding models, which can be used to benchmark text-to-3D generation methods, select generated 3D assets with better quality and help optimize text-to-3D models,\textit{etc.}} This paper makes the following contributions:
\begin{itemize}
    \item We construct so far the largest text-to-3D assets quality assessment database, named AIGC-T23DAQA database, and to the best of our knowledge, this is the first work that tries to study human preferences for AI-based text-to-3D assets from multiple perspectives.
    
    \item We propose a novel projection-based evaluator for better text-to-3D asset quality assessment, termed T23DAQA model, which leverages a 3D encoder, two 2D encoders, and multi-modality foundation models to extract features encompassing 3D shape, texture, and text-asset correspondence to predict human preference scores.
    
    \item Comprehensive experimental results demonstrate that our proposed method surpasses existing state-of-the-art NR-IQA, NR-VQA, NR-MQA, NR-PCQA, LMMQA, T2IQA, T2VQA models, and text-image alignment methods, affirming its efficacy in measuring the perceptual quality of text-to-3D assets. Furthermore, the ablation experiments validate the effectiveness of the proposed module.
    
\end{itemize}

\section{Related Work}
\subsection{Text-to-3D Asset Generation}
In recent years, many 3D asset generation methods have been proposed, drawing inspiration from advancements in AI-based 2D image generation works. Early explorations in 3D generation \cite{henzler2019escaping} have leveraged generative adversarial network (GAN) algorithms, such as 3DGAN, to produce 3D models from probability space. The seminal work DreamFusion \cite{poole2022dreamfusion} have pioneered the utilization of pre-trained 2D text-to-image models for text-to-3D transformation via differentiable rendering. Their key methodology, score distillation sampling (SDS), involves uniformly sampling from the parameter space of pre-trained diffusion models to obtain gradients aligned with given text prompts. Building upon this foundation, Magic3D \cite{lin2023magic3d} have further enhanced the quality and efficiency of 3D asset generation through a two-step approach. Prolificdreamer \cite{wang2024prolificdreamer}, SJC\cite{wang2023score}, LatentNerf \cite{metzer2023latent} and TextMesh \cite{tsalicoglou2023textmesh} have optimized 3D asset generation by improving the representation of 3D assets and improving SDS. These works generally employ volunteers to conduct pairwise comparisons of results from different methods to ascertain the visual quality of generated 3D asset, underscoring the pressing need for a quality assessment algorithm tailored to generated 3D asset.

\begin{table*}[t]
\caption{Summary of the existing AIGC Quality Assessment databases and AIGC-T23DAQA database. The numbers in parentheses of score type represent the dimensions of the subjective experimental annotations.\label{tab:summary}}
\centering
\begin{tabular}{c|c|c|c|c|c|c|c|c}
\toprule
Type                           & \multicolumn{1}{c|}{Dataset} & \multicolumn{1}{c|}{Contents}         & \multicolumn{1}{c|}{Prompts} & \multicolumn{1}{c|}{Models} & \multicolumn{1}{c|}{Annotators} & \multicolumn{1}{c|}{Ratings} & \multicolumn{1}{c|}{Score type}   & \multicolumn{1}{c}{Public Available} \\ \hline
\multirow{8}{*}{Text-To-Image} & Pick-a-pic\cite{kirstain2024pick}        & 500,000         & 35,000  & 3  & -  & 500,000   & Preference         & $\checkmark$ \\
                               & HPS\cite{wu2023human}                    & 98,807         & 25,205  & 1  & -  & 98,807    & Preference         & $\checkmark$    \\
                               & ImageReward\cite{xu2024imagereward}      & 136,892 & 8,878   & 1  & -  & 136,892   & Seven Point Likert & $\checkmark$ \\
                               & AGIQA-1K\cite{zhang2023perceptual}       & 10,80                & 540     & 2  & 22 & 23,760    & MOS                & $\checkmark$   \\
                               & AGIQA-3K\cite{10262331}                  & 2,982             & 497     & 6  & 21 & 125,244   & MOS(2)             & $\checkmark$   \\
                               & AGIQA-20K\cite{li2024aigiqa}             & 20,000                & 20,000  & 15 & 21 & 420,000   & MOS                & $\checkmark$  \\
                               & AIGCIQA2023\cite{wang2023aigciqa2023}    & 2,400            & 100     & 6  & 28 & 201,600   & MOS(3)             & $\checkmark$  \\ 
                               & AIGCOIQA2024\cite{yang2024aigcoiqa2024}  & 300            & 25     & 5  & 20 & 18,000   & MOS(3)             & $\checkmark$  \\ \hline
\multirow{5}{*}{Text-To-Video} & Chivileva's\cite{chivileva2023measuring} & 1,005             & 201     & 5  & 24 & 48,240    & MOS(2)             & $\checkmark$     \\
                               & EvalCrafter\cite{liu2023evalcrafter}     & 3,500             & 700     & 7  & 3  & 73,500    & MOS(5)             & $\checkmark$  \\
                               & Vbench\cite{huang2023vbench}             & 6,984         & 1,746   & 4  & -  & 209,520   & Preference         & $\checkmark$  \\
                               & FETV\cite{liu2023fetv}                   & 2,476             & 619     & 3  & 3  & 11,142    & MOS(2)             & $\checkmark$  \\
                               & T2VQA-DB\cite{kou2024subjective}         & 1,000                & 1,000   & 9  & 27 & 27,000    & MOS                & $\checkmark$  \\ \hline
Text-To-3D                     & Ours                                     & 969             & 170     & 6  & 17 & 49,419    & MOS(3)             & $\checkmark$   \\
\bottomrule
\end{tabular}
\vspace{-0.5cm}
\end{table*}
\vspace{-15pt}
\subsection{3D Quality Assessment}
3D asset quality assessment can be used to choose or optimize 3D assets, and contribute to VR \cite{duan2024quick, duan2023attentive, duan2018perceptual} and AR \cite{duan2022augmented, duan2022saliency} applications. Currently, most 3D asset quality assessment studies mainly research the mesh quality assessment (MQA) and point cloud quality assessment (PCQA) problems, as mesh and point cloud formats represent common structures of 3D models. According to the quality feature extraction methods, the MQA methods can be divided into two main categories, including: model-based approaches and projection-based approaches. Model-based methods \cite{m1, tian-color} typically compute local features at the vertex level and global color features from texture images, subsequently aggregating these features into the quality score. However, projection-based methods \cite{zhang2023gms} need to first generate projection images from the mesh, then utilize mature 2D IQA or 3D video quality assessment (VQA) tools to predict mesh quality scores. Similarly, PCQA methods can also be categorized into model-based and projection-based methods. However, due to the discrepancy in data storage between point clouds and meshes, model-based PCQA methods \cite{tian2017geometric, torlig2018novel} typically extract geometry features from point-wise gradient vector distances and color features from point-wise color attributes. Projection-based PCQA methods \cite{zhang2022treating} generally follow projection-based MQA methods,  which extracts features from the projected images of point clouds by 2D IQA and 3D VQA tools to predict quality scores.

Model-based methods does not exist information loss during evaluating but demand considerable computational resources due to the complexity of high-fidelity point clouds or meshes, while projection-based approaches relying on mature 2D IQA or 3D VQA tools have lower computational complexity, but the performance may be influenced by the selection of viewpoints. To mitigate the variability inherent in viewpoint selection, several studies \cite{zhang2022treating, fan2022no} advocate for employing multiple projections, significantly enhancing accuracy compared to single-projection approaches. Compared to traditional 3D quality assessment tasks and methods \cite{zhang2022no, liu2021pqa, chai2024plain}, our proposed database and method focus on text-to-3d asset, which is more potential and challenge.

\vspace{-15pt}
\subsection{AIGC Image and Video Quality Assessment}
With the success of diffusion models in image generation tasks, numerous text-to-image methods have emerged. Concurrently, to evaluate the quality of AI-generated images (AIGI), several AGI databases and AI-generated image quality assessment (IQA) methods have surfaced. These databases can be classified into two main types including coarse-grained and fine-grained. Coarse-grained databases such as HPS\cite{wu2023human} and Pick-A-Pic \cite{kirstain2024pick} generally gather paired image comparison results or series image selection results for images generated by Stable Diffusion or other text-to-image models, as subjective evaluation results. In contrast, fine-grained databases like AGIQA-20K \cite{li2024aigiqa} and AIGCIQA2023 \cite{wang2023aigciqa2023} generally conducts subjective quality rating experiments from multiple perspectives to evaluate human preferences for AIGIs. For objective AIGI quality assessment, IS \cite{gulrajani2017improved} and FID \cite{heusel2017gans} have long been adopted to evaluate the fidelity of a collection of generated images. Recently, numerous specialized algorithms for evaluating AIGIs have emerged \cite{xu2024imagereward, wu2023human}. These algorithms typically leverage contrastive language-image pre-training (CLIP) \cite{radford2021learning} to extract text-image features and utilize classical classification backbone networks to extract image perception features\cite{wang2024understanding}, which are then fused to predict preference scores.

Recently, OpenAI's video generation model Sora has demonstrated the ability to generate one-minute high-fidelity videos, drawing public attention to the task of text-generated videos. Recently, some quality assessment studies for AI-generated videos (AIGV) have also been conducted. Chivileva et al. \cite{chivileva2023measuring} have proposed a dataset comprising 1,005 videos generated by 5 text-to-video models, with quality assessment performed by 24 annotators to provide subjective scores. Similarly, Kou et al. \cite{kou2024subjective} have established the expansive text-to-video quality assessment database (T2VQA-DB), consisting of 10,000 videos generated by 9 different text-to-video models, each accompanied by its corresponding MOS. Text-to-video quality assessment algorithms typically combine NR-VQA and 2D AIGIQA methods to predict text-to-video quality. An overview of current AIGC quality assessment databases have been give in Table \ref{tab:summary}.


\section{Database Construction and Analysis}
In this section, we will describe the database construction and analysis in detail.
\begin{figure}[b]
    \centering
    \includegraphics[width = 0.48\textwidth]{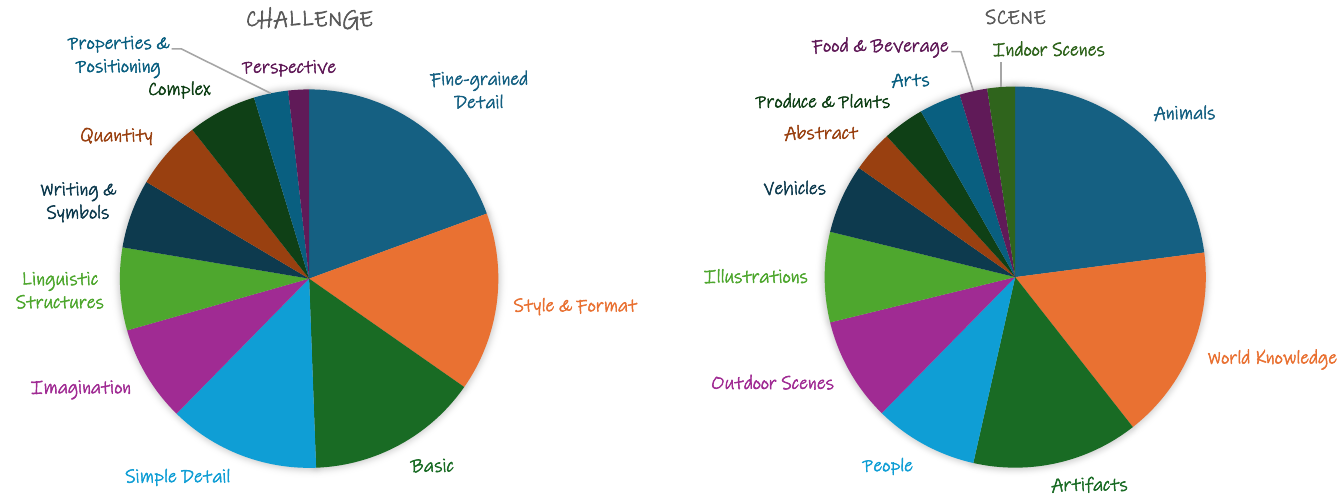}
    \caption{The Pie Chart of our used Prompt, which contains 11 challenge categories and 12 scene categories.}
    \label{fig:pie}
\end{figure}
\begin{figure*}[t]
    \centering
    \subfigure[3D assets generated by the prompt: ``a harp without any strings'']{\begin{minipage}[t]{\linewidth}
                \centering
                \includegraphics[width = 0.98\linewidth]{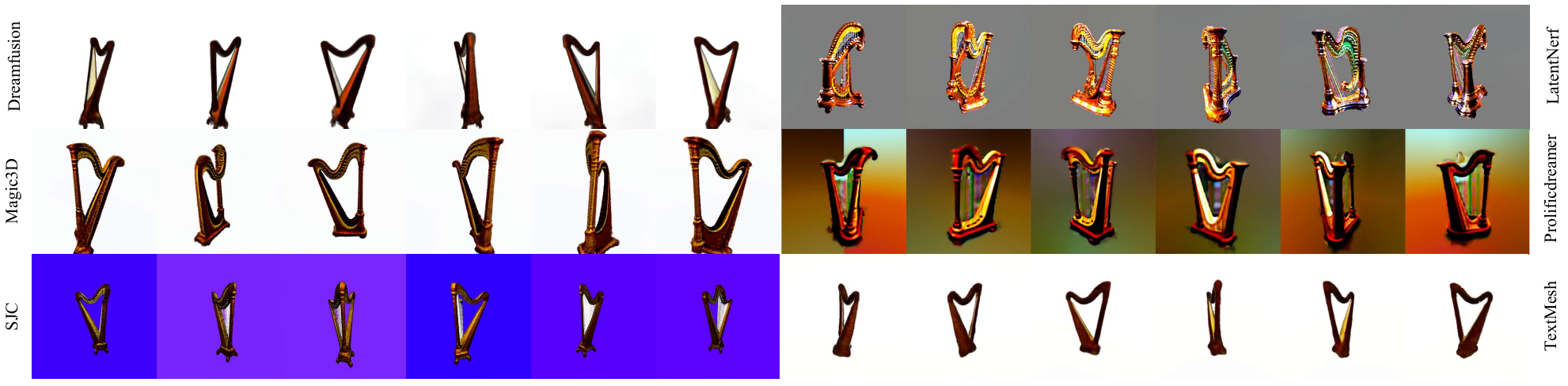}
                \end{minipage}}
                
    \subfigure[3D assets generated by the prompt: ``a pair of brown suede shoes'']{\begin{minipage}[t]{\linewidth}
                \centering
                \includegraphics[width = 0.98\linewidth]{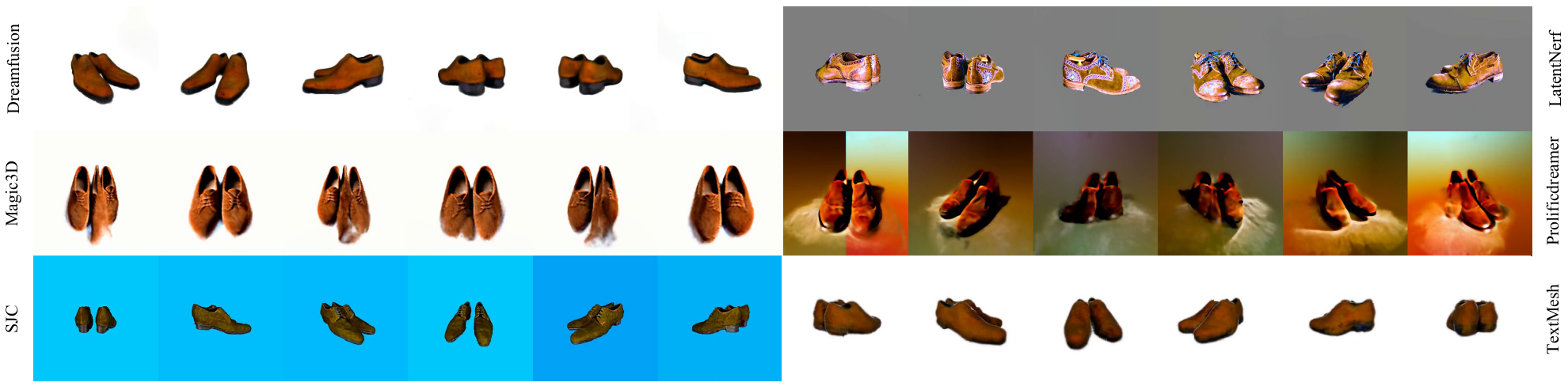}
                \end{minipage}}
                
    \caption{Sample 3D assets from the AIGC-T23DAQA database, generated by Dreamfusion \cite{poole2022dreamfusion}, LatentNerf \cite{metzer2023latent}; Magic3D \cite{lin2023magic3d}, Prolificdreamer \cite{wang2024prolificdreamer}; SJC\cite{wang2023score}, TextMesh \cite{tsalicoglou2023textmesh} with the same input prompt respectively. (a) 3D assets generated by the prompt ``a harp without any strings''. (b) 3D assets generated by the prompt ``a pair of brown suede shoes''. This clearly shows that the visual quality of assets generated by different models varies greatly.}
    \label{fig:gallery}
    \vspace{-0.5cm}
\end{figure*}
\vspace{-15pt}
\subsection{Prompt Selection}
Compared to AIGC IQA and VQA databases, constructing text-to-3D asset quality assessment database mainly faces two difficulties: 1) The process of generating 3D asset from text is currently time-consuming, typically requiring 1 to 6 hours to generate one 3D asset. 2) The subjective experiment for evaluating generated 3D asset is also time-consuming, since subjects need to observe from whole directions and assess from multiple perspectives. Therefore, our constructed database is a enormous contribution to the field. First of all, meticulous prompts selection is important for text-to-3D asset quality assessment database construction. The selected prompts need to cover a wide range of real user inputs with a relatively small pool. PartiPrompts \cite{yu2022scaling} comprises 1600 varied English prompts designed to comprehensively assess and test the limits of text-to-image synthesis models. Following previous research \cite{wang2023aigciqa2023} we extracted 170 prompts from PartiPrompts, spanning 11 challenge categories and 12 scene categories. The distribution of selected scene and challenge categories is depicted in the pie chart of Fig. \ref{fig:pie}, which manifests that the prompts in our dataset exhibit a high level of scene diversity and encompass a broad spectrum of challenges.
\vspace{-15pt}
\subsection{3D Asset Generation}

To ensure asset diversity, AIGC-T23DAQA database contain six representative text-to-3D asset generation models. These current models typically comprise a 2D image generation module and a 3D asset representation module. When compared to other generation models, the diffusion model delivers exceptional results, establishing itself as the preferred foundational module for generating 2D images within these methodologies. For the 3D asset representation module, a variety of approaches are employed, including NeRF, Instatn-ngp, \textit{etc}. Dreamfusion \cite{poole2022dreamfusion} utilizes mip-NeRF 360 for 3D asset representation, while LatentNerf \cite{metzer2023latent} opts for vanilla NeRF. SJC \cite{wang2023score} employs voxel radiance fields to represent 3D asset, thereby enhancing the speed of the generation process. Conversely, TextMesh  \cite{tsalicoglou2023textmesh}, Magic3D \cite{lin2023magic3d}, and Prolificdreamer \cite{wang2024prolificdreamer} adopt a coarse-to-fine strategy. They commence with coarse 3D asset representations, using vanilla NeRF and Instatn-ngp, respectively, and subsequently refine the differentiable mesh into a fine representation. 
\begin{figure}[!t]
    \centering
    \subfigure[Prompt: ``an ostrich''.]{\begin{minipage}[t]{0.48\linewidth}
                \centering
                \includegraphics[width = 0.94\linewidth]{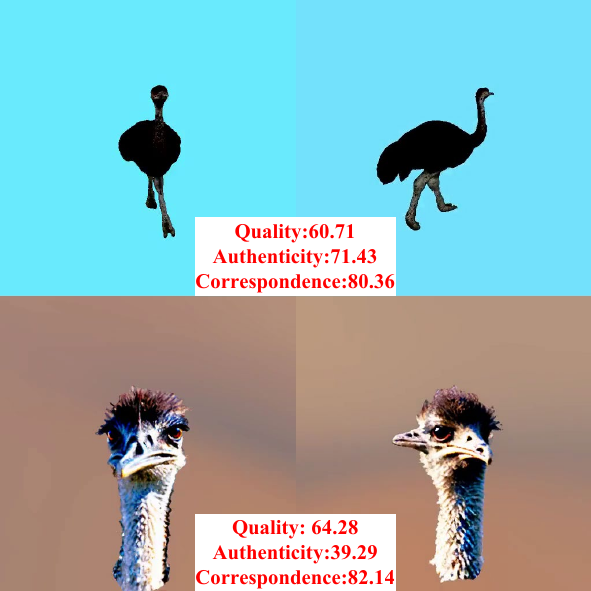}
                \end{minipage}}
    \subfigure[Prompt: ``a comic about a boy and a tiger''.]{\begin{minipage}[t]{0.48\linewidth}
                \centering
                \includegraphics[width = 0.94\linewidth]{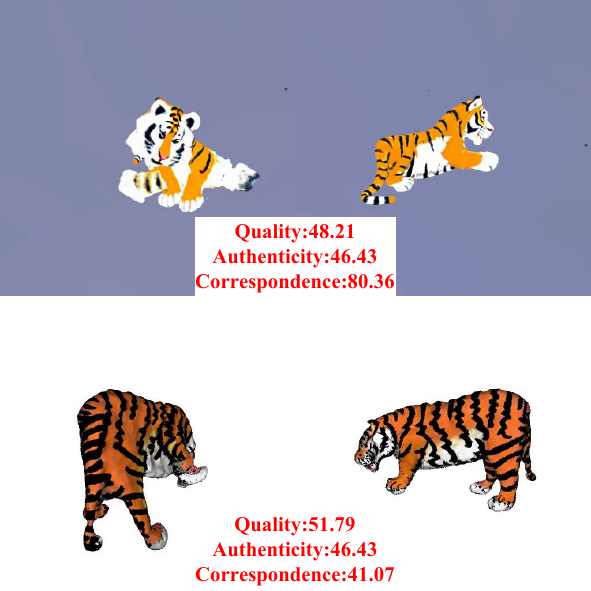}
                \end{minipage}}
                
    \subfigure[Prompt: ``a fish without eyes''.]{\begin{minipage}[t]{0.48\linewidth}
                \centering
                \includegraphics[width = 0.94\linewidth]{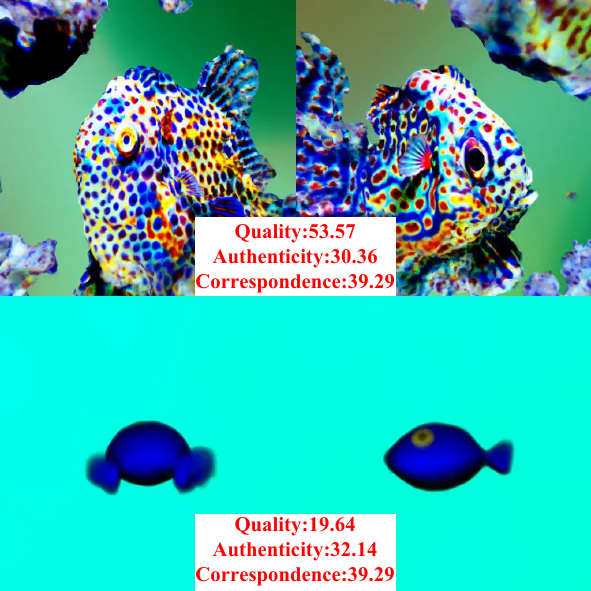}
                \end{minipage}}
    \subfigure[Prompt: ``a large present with a red ribbon to the left of a Christmas tree''.]{\begin{minipage}[t]{0.48\linewidth}
                \centering
                \includegraphics[width = 0.94\linewidth]{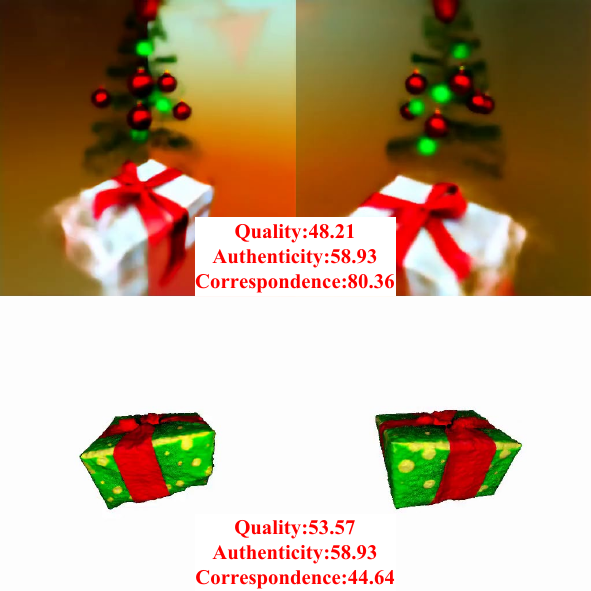}
                \end{minipage}}
                
    \subfigure[Prompt: ``a robot cooking''.]{\begin{minipage}[t]{0.48\linewidth}
                \centering
                \includegraphics[width = 0.94\linewidth]{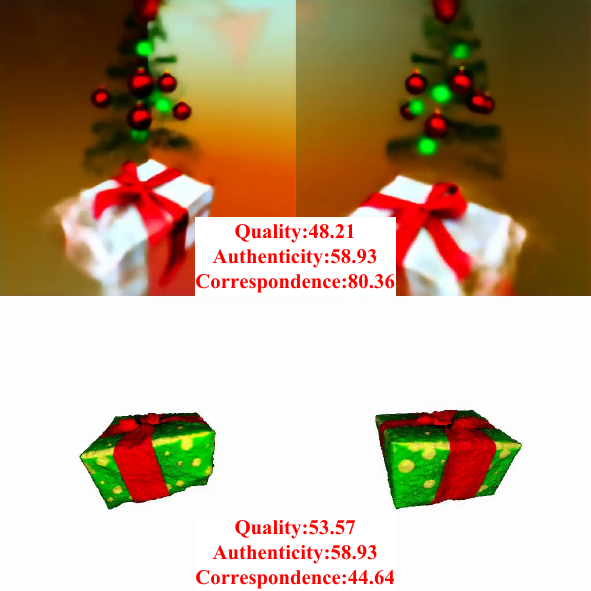}
                \end{minipage}}
    \subfigure[Prompt: ``a bundle of blue and yellow flowers in a vase''.]{\begin{minipage}[t]{0.48\linewidth}
                \centering
                \includegraphics[width = 0.94\linewidth]{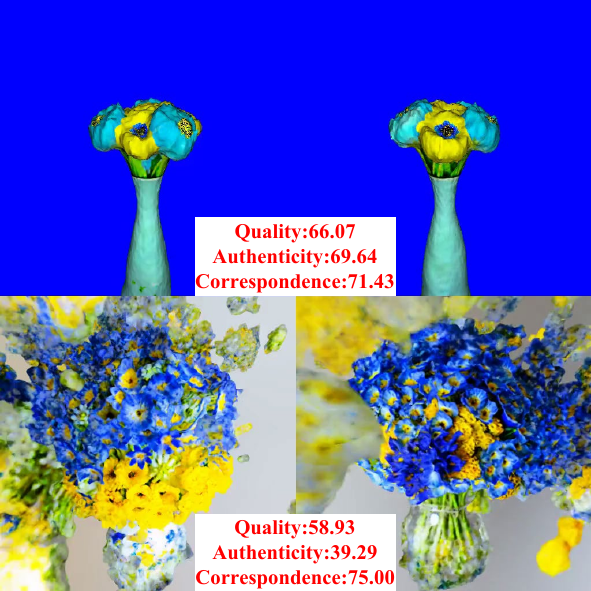}
                \end{minipage}}
    \caption{Illustration of the differences between the three dimensions of quality ,authenticity, and text-3D correspondence. In each subfigure, the images in the top row are significantly better than the that in bottom row in terms of two perspectives, while similar or worse in terms of another perspective. (a) and (b) show examples that the authenticity and correspondence scores of the top images are higher, while the quality is similar. (c) and (d) show examples that the quality and correspondence scores of the top images are higher, while the authenticity is similar or lower. (e) and (f) show examples that the quality and authenticity scores of the top images are higher, while the correspondence is similar or lower. }
    \label{fig:xx}
    \vspace{-0.5cm}
\end{figure}
The generation process of text-to-3D asset was executed using open-source code \cite{threestudio2023} with default weights and configurations, resulting in a collection of 1020 instances (170 prompts × 6 models) of text-to-3D assets. Some examples of the 3D assets generated by the six text-to-3D asset generation models are illustrated in Fig.  \ref{fig:gallery}. Subsequently, we discarded 51 instances of failed asset generation, defined as cases where the entire spatial domain remained empty after-generation. Due to computational constraints, it is hard to render a generated 3D asset in real-time and evaluate it. Thus, we followed the method used in \cite{zhang2023eep} and projected the 3D asset into videos then conducted evaluation. This manipulation yielded 969 360-degree surround projection videos centered on the generated text-to-3D asset. Each video consists of comprised 120 frames with a resolution of 512 × 512 pixels and cumulative a total duration of 4 seconds. These projection videos were used for the subsequent subjective experiment.

\vspace{-10pt}
\subsection{Subjective Experiment}
\begin{figure}[!t]
    \centering
    \subfigure[Quality]{\begin{minipage}[t]{\linewidth}
                \centering
                \includegraphics[width = 0.98\linewidth]{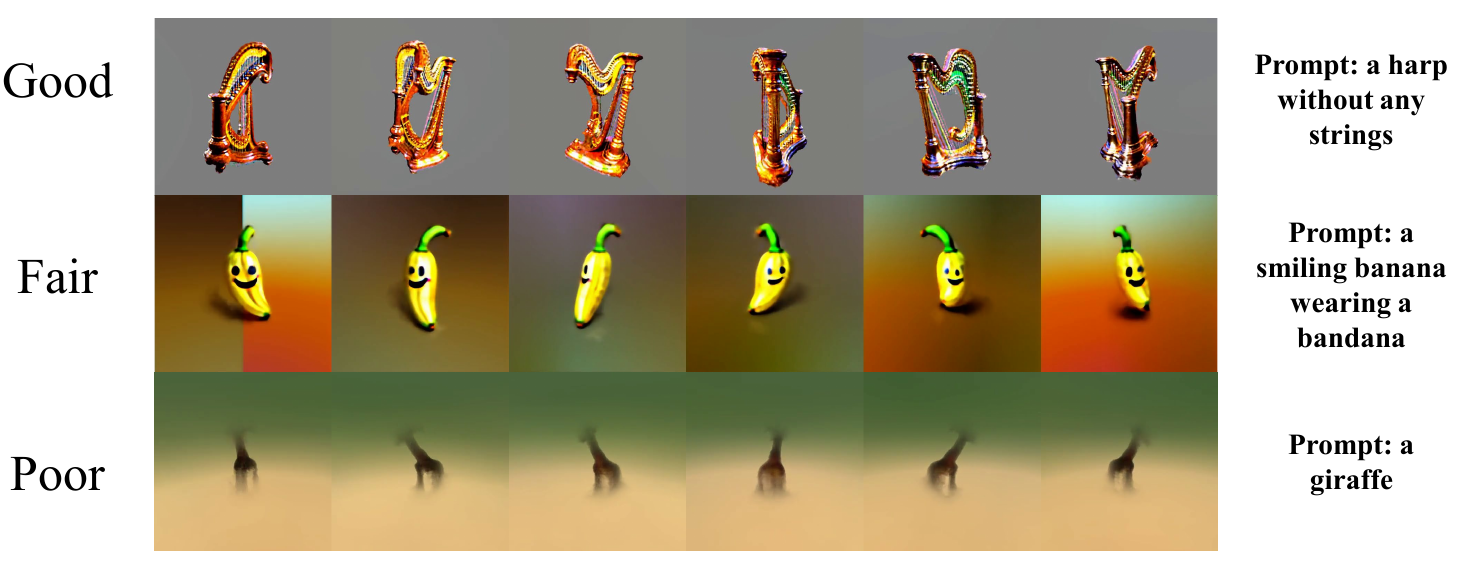}
                \end{minipage}}
                
    \subfigure[Authenticity]{\begin{minipage}[t]{\linewidth}
                \centering
                \includegraphics[width = 0.98\linewidth]{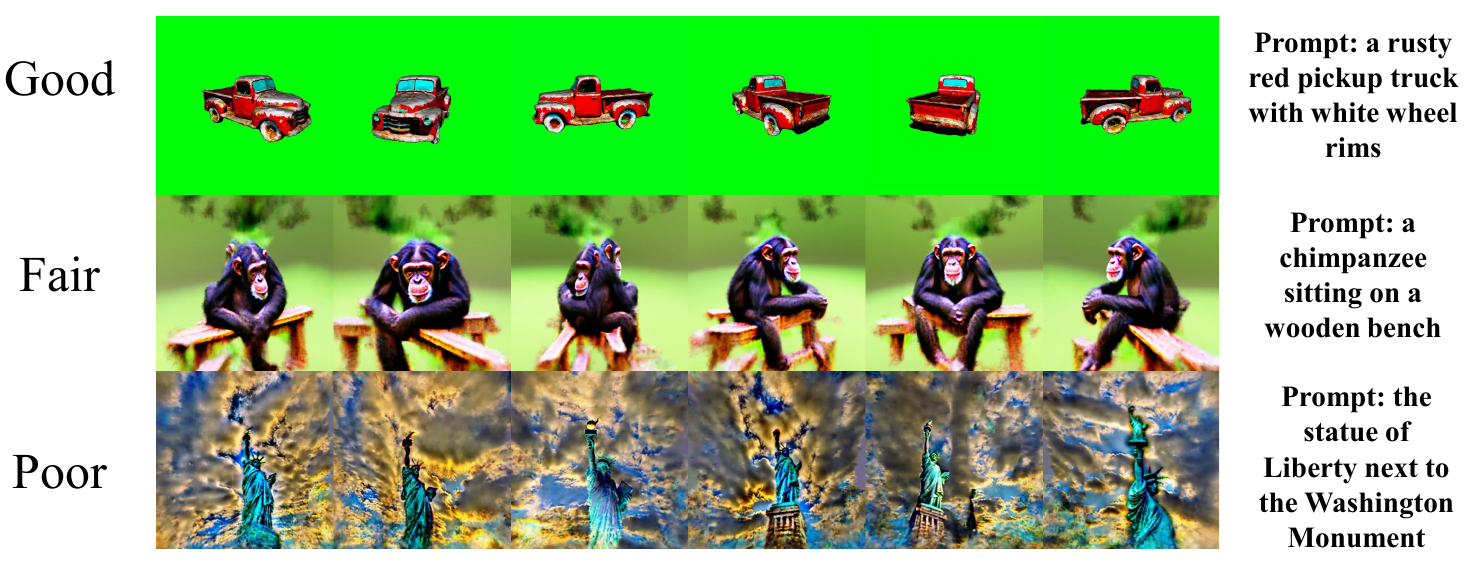}
                \end{minipage}}

    \subfigure[Correspondence]{\begin{minipage}[t]{\linewidth}
                \centering
                \includegraphics[width = 0.98\linewidth]{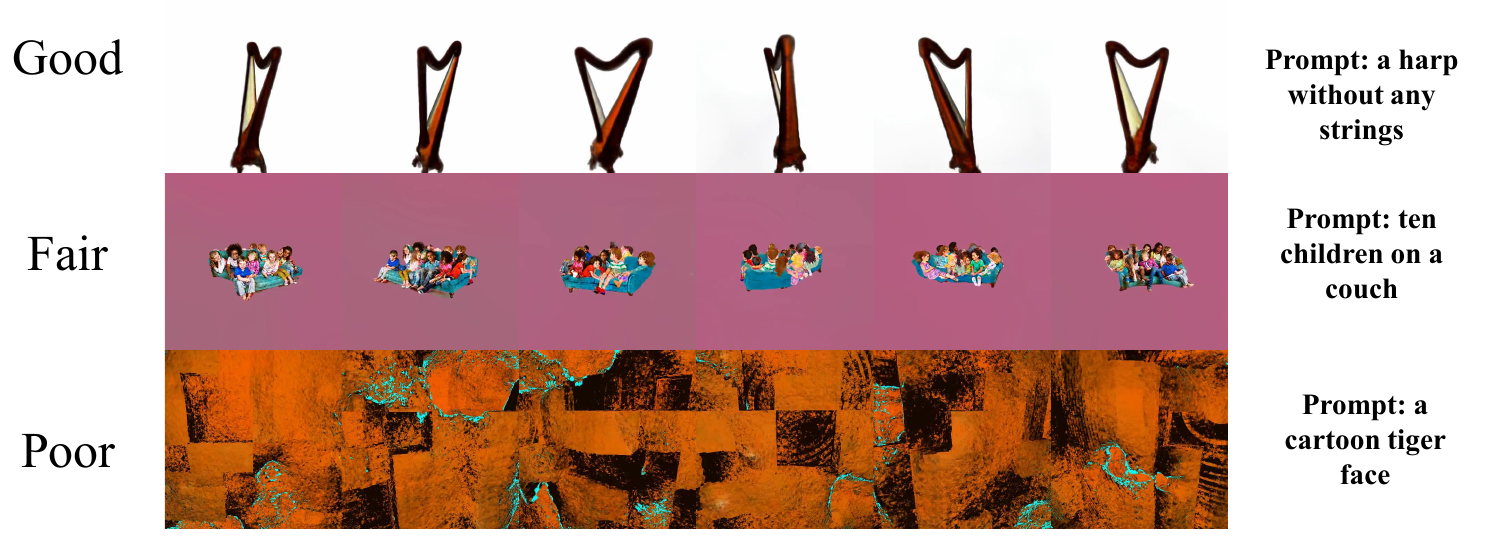}
                \end{minipage}}
    \caption{Illustration of the text-to-3d assets from the perspectives of quality, authenticity, and text-asset correspondence. The examples of good, fair, and poor quality are depicted in the first to third rows of (a). The examples illustrating good, fair, and poor authenticity are displayed in the first to third rows of (b). (c) showcases examples of good, fair, and poor correspondence generated by prompts ``a harp without any strings'', ``a knight holding a long sword'', and ``A cartoon tiger face''. }
    \label{fig:ap}
    \vspace{-0.4cm}
\end{figure}
To collect human visual preferences for text-to-3D assets, we further conducted a subjective evaluation experiment. As highlighted in prior AI generated asset quality assessment studies \cite{wang2023aigciqa2023, yang2024aigcoiqa2024}, the degradations of AI generated asset are significantly different from human captured or created asset, which need to be evaluated from multiple perception perspectives. \textcolor{black}{Based on traditional 3D quality assessment, which evaluates texture, color, and other visual quality attributes of the 3D asset, we selected the ``quality'' dimension for evaluation. Similar to AI-generated image and video quality assessment, in addition to assessing the visual quality of the 3D asset, we also need to evaluate its authenticity and correspondence to the text prompt. Therefore, we selected the dimensions of ``authenticity'' and ``correspondence''.} Hence, in this paper, we propose to evaluate human visual preferences for text-to-3D assets from three perspectives, including quality, authenticity, and text-asset correspondence. Fig. \ref{fig:xx} shows the differences between the selected three dimensions, which further manifests the importance, and significance of evaluating text-to-3D assets from multiple perspectives.
\textcolor{black}{Before each subject conducts the subjective experiment, we give a detailed instruction to subjects, which includes explaining to the subject the differences between ``quality'', ``authenticity'' and ``correspondence'' and showing examples of different degrees of each dimension. The ``quality" is the visual quality attribution of 3D asset including texture, color, integrity, etc, while the ``authenticity" refers to whether the 3D asset is consistent with the real world that the subject knows. The ``correspondence" is the alignment between the 3D asset and the input prompt text.} Then, participants were instructed to give their preference scores of text-to-3D assets based on the surrounding 360-degree projection videos. The first dimension for evaluating text-to-3D asset is ``quality'', which mainly evaluates the perception attributes including texture, color, integrity, details \textit{etc.}, analogous to traditional 3D models. Fig. \ref{fig:ap} (a) shows examples of the generated 3D asset with different ``quality'' levels. The second dimension for evaluating text-to-3D asset is ``authenticity'', which evaluates the perception attributes including unrealistic textures, shapes, \textit{etc.} It should be noted that compared to the authenticity attribute generally used in AIGC IQA, the degradation of the authenticity attribute for generated text-to-3D asset generally comes from the unrealistic or inconsistent multiple views. Fig. \ref{fig:ap} (b) shows examples of the generated 3D asset with different ``authenticity'' levels. Similar to AIGC IQA, and AIGC VQA methodologies, the correspondence between text, and 3D asset serves as another critical criterion in assessing text-to-3D asset quality, referred to as ``text-3D asset correspondence''. Fig. \ref{fig:ap} (c) shows examples of the generated 3D asset with different ``correspondence'' levels.
\begin{figure}[t]
    \centering
    \includegraphics[width = 0.48\textwidth]{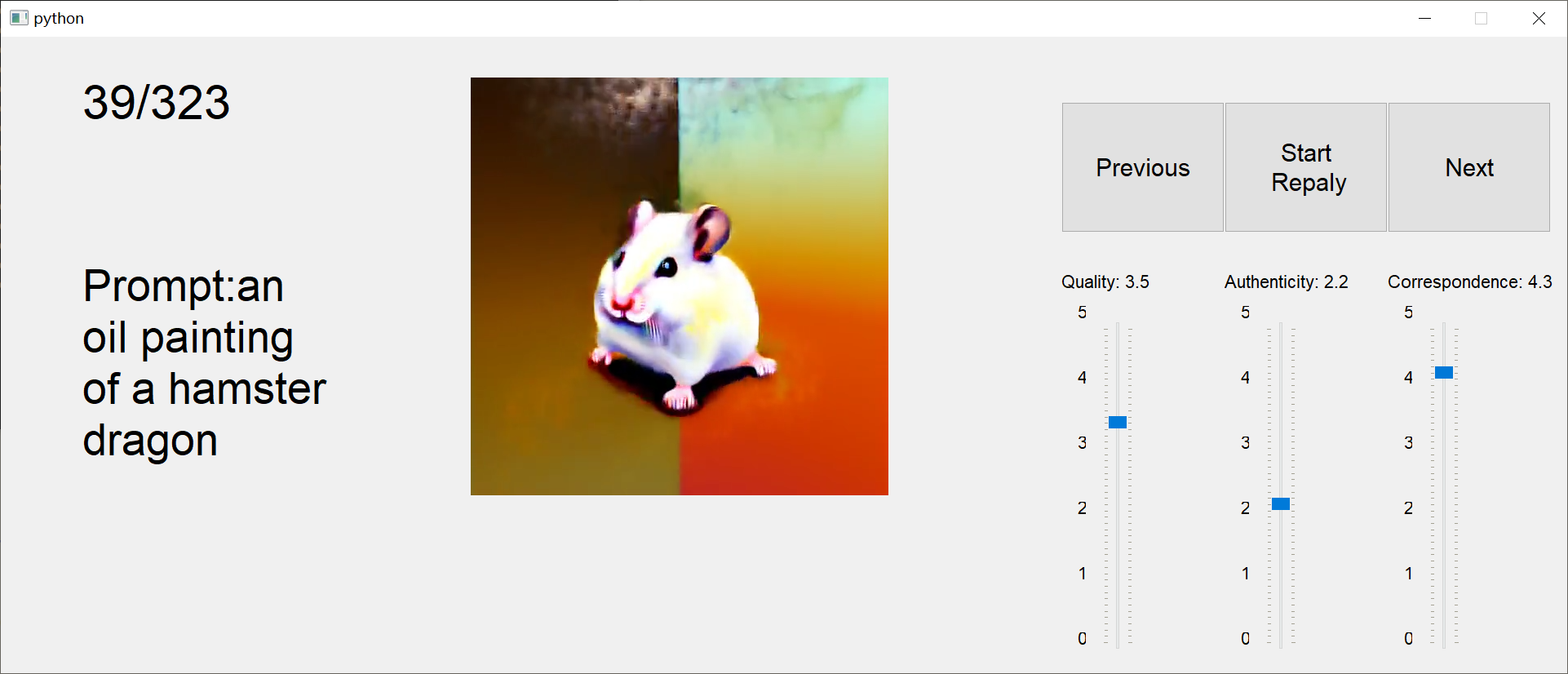}
    \caption{The illustration of the subjective assessment interface. The subject can evaluate their preferences of the text-to-3D assets, and record the quality, authenticity, correspondence scores with the scroll bars on the right.}
    \label{fig:gui}
    \vspace{-0.5cm}
\end{figure}

\begin{figure*}[t]
    \centering
    \subfigure[Quality MOS distribution.]{\begin{minipage}[t]{0.32\linewidth}
                \centering
                \includegraphics[width = 0.94\linewidth]{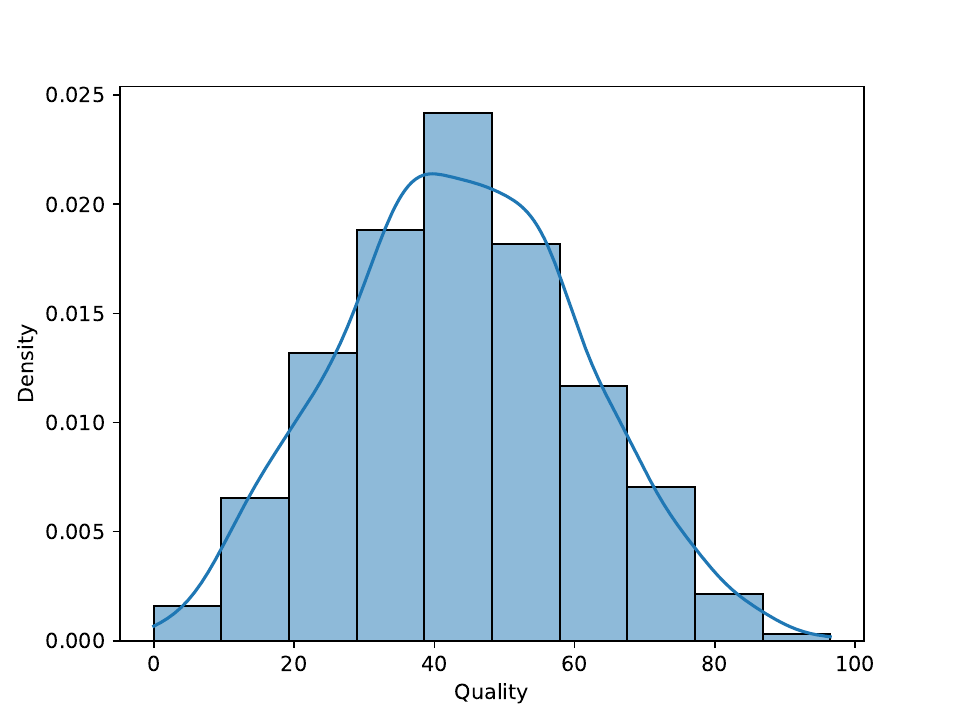}
                \end{minipage}}
    \subfigure[Authenticity MOS distribution.]{\begin{minipage}[t]{0.32\linewidth}
                \centering
                \includegraphics[width = 0.94\linewidth]{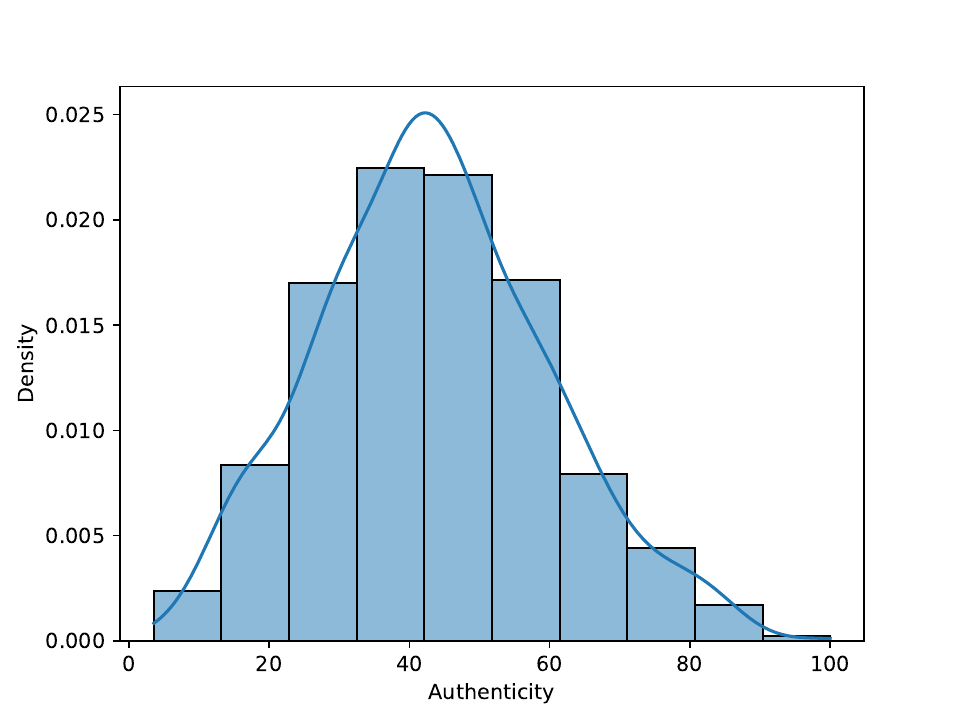}
                \end{minipage}}
    \subfigure[Correspondence MOS distribution.]{\begin{minipage}[t]{0.32\linewidth}
                \centering
                \includegraphics[width = 0.94\linewidth]{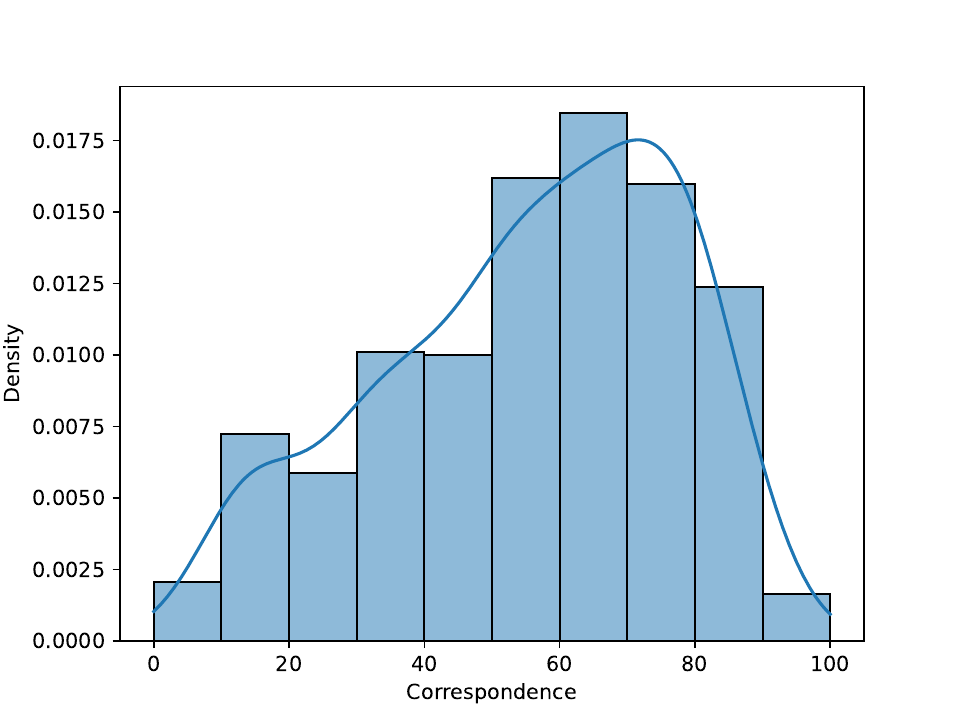}
                \end{minipage}}
    \caption{Distributions of the MOSs from the perspectives of quality, authenticity, and correspondence, respectively. These distributions exhibiting proposed T23DAQA database cover a wide range in terms of all perspectives. }
    \label{fig:mos}
    \vspace{-0.5cm}
\end{figure*}

We conducted the subjective experiment following the guidance in ITU-R BT.500-13 \cite{other:itu}. The experimental environment was arranged to simulate a typical indoor home setting with standard lighting conditions. The projection videos of text-to-3D asset, accompanied by the corresponding prompts, were presented randomly on a monitor with a resolution of $1920 \times 1080$. The interface, depicted in Fig. \ref{fig:gui}, facilitated viewer interaction, enabling navigation through previous, next, and replay options for the projection videos of the generated 3D asset. Additionally, three sliders ranging from 0 to 5, with a minimum interval of 0.1, were provided for participants to assign scores for quality, authenticity, and correspondence. 17 subjects (10 males and 7 females) participated in the subjective experiment, all possessing normal or corrected-to-normal vision. Each participant received detailed experimental instructions prior to engaging in the subjective evaluation. We divided the conversation of each participant in the subjective experiment into three subsets. For each participant, the database were randomly divided into three subsets, which are used in three subjective tests respectively. Each test lasted around one hour, followed by a 10-20 minutes break in between, and then the next test was performed.
\vspace{-15pt}
\subsection{Data Processing}

We followed the instructions of ITU \cite{other:itu} to conduct the outlier detection and subject rejection. \textcolor{black}{Specifically, for each evaluation dimension, we calculate the kurtosis of the raw subjective quality ratings for each generated 3D asset to determine whether the data follows a Gaussian or non-Gaussian distribution. For Gaussian distributions, a raw score is considered an outlier if it lies more than 2 standard deviations (std) from the mean. For non-Gaussian distributions, a score is deemed as an outlier if it is more than $\sqrt{20}$ standard deviations from the mean. Any subject whose evaluations exceed a 3\% outlier rate in any dimension is excluded from the analysis.} As a result, no subjects were rejected and the rejection ratio is 3\% for all ratings. Subsequently, we converted the raw ratings of the remaining valid subjective scores into Z-scores, which were then linearly scaled to the range of $\left [ 0,100 \right ]$. The final MOS is computed as follows:
\begin{equation}
z_{ij} = \frac{m_{ij}-\mu_{i}}{\sigma_{i}}, \quad z_{ij}^{'} = \frac{100\times(z_{ij} + 3)}{6}
\end{equation}

\begin{equation}
\text{MOS}_{j} = \frac{1}{N} z_{ij}^{'}
\end{equation}
where $m_{ij}$ is the subjective score given by the $i$-th subject to the $j$-th text-to-3D asset, $\mu_{i}$ and $\sigma_{i}$ is the mean score and the standard deviation given by the $i$-th subject respectively, $N$ is the total number of subjects. 
\vspace{-15pt}
\subsection{Subjective Data Analysis}
Although a large number of text-to-3D asset generation models have been developed in recent years, the corresponding works that specifically analyze and compare their generation performance are lacking. Considering that the generation quality of the text-to-3D asset is influenced by multiple factors such as prompts, algorithms, \textit{etc}, which leads to diverse perceptual quality and affects the user experience, based on the established AIGC-T23DAQA database, we conduct an in-depth analysis for the collected MOSs from multiple perspectives as follows.

Fig. \ref{fig:mos} demonstrates the distribution of MOS values obtained from subjective experiments. It can be observed that the correspondence distribution surpasses both the quality and authenticity distributions, suggesting that the current generation models learn more towards correspondence while ignoring the quality and authenticity attributes. The reason for this phenomenon is that the current T23DA method utilizes text-to-image models to constrain the correspondence between images rendered from different perspectives and text. These text-to-image models are trained on a large number of text-image pairs and perform well in text-image correspondence, ensuring good correspondence between generated 3D asset and text; However, the text-generated image model cannot guarantee the geometric texture consistency of three-dimensional objects from different perspectives, resulting in the strange geometric shapes and floaters in generated 3D asset. As a result, the quality and authenticity of the generated 3D assets are poorer than those of correspondence. To enhance the overall user preferences in the future, it is more important to improve the quality and authenticity attributes for the generated 3D assets.

\begin{figure}[ht]
    \centering

    \subfigure[]{\begin{minipage}[t]{0.9\linewidth}
                \centering
                \includegraphics[width = 0.93\linewidth]{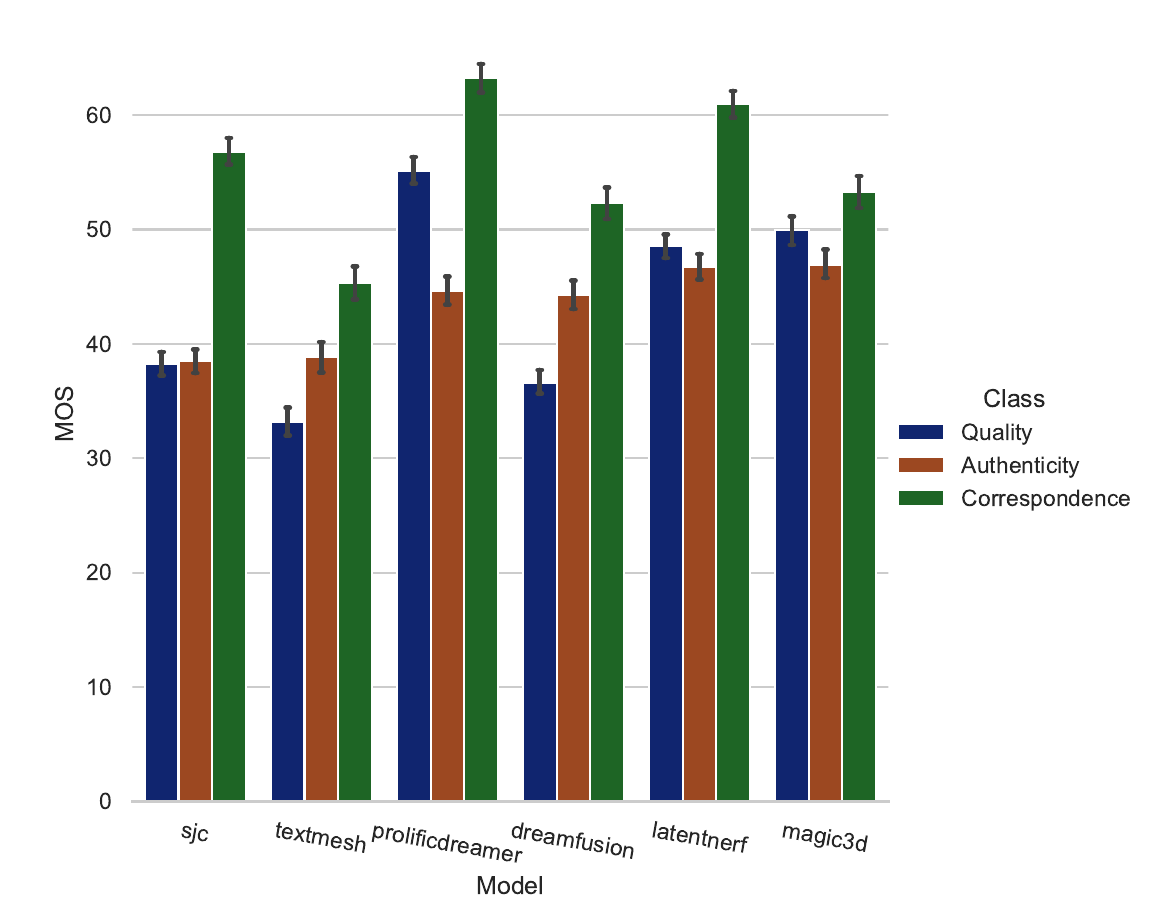}
                \end{minipage}}

    \subfigure[]{\begin{minipage}[t]{0.9\linewidth}
                \centering
                \includegraphics[width = 0.93\linewidth]{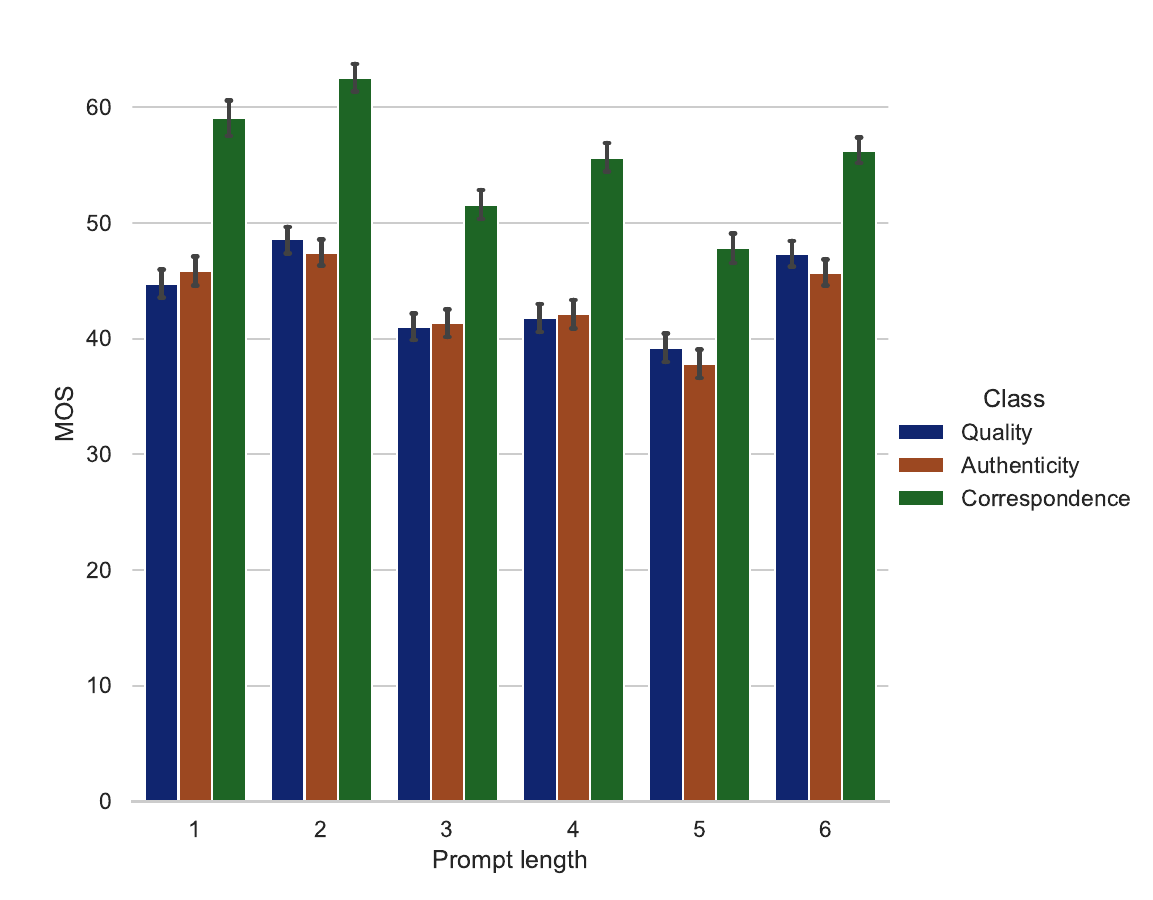}
                \end{minipage}}
                
    \caption{Illustration of the impact of different models and prompt lengths on the perceptual quality of T23DAs respectively. (a) shows the subjective quality, authenticity, and correspondence score of T23DAs with different methods including Dreamfusion, LatentNerf, Magic3D, Prolificdreamer, SJC, and TextMesh respectively. (b) shows the subjective quality, authenticity, and correspondence score of T23DAs with different prompt lengths. Prompt length is divided into six intervals on average, with 1-6 on the x-axis representing interval numbers from short to long.}
    \label{fig:moscompare}
    \vspace{-0.6cm}
\end{figure}

Fig. \ref{fig:moscompare} (a) compares the human preference MOSs for different models, including Dreamfusion \cite{poole2022dreamfusion}, LatentNerf \cite{metzer2023latent}; Magic3D \cite{lin2023magic3d}, Prolificdreamer \cite{wang2024prolificdreamer}; SJC\cite{wang2023score}, TextMesh \cite{tsalicoglou2023textmesh}. Fig. \ref{fig:moscompare} (b) compares the human preference MOSs for different prompt length. Prompt length is divided into six intervals on average, with 1-6 on the x-axis representing interval numbers from short to long. We can find from it that: 1) The 3D assets generated by different text-to-3D generation models have significantly different perceptual preferences, and even with the same input prompt, the quality, authenticity, and correspondence vary greatly across different text-to-3D asset methods. Models including Prolificdreamer \cite{wang2024prolificdreamer}, Magic3D \cite{lin2023magic3d}, and Prolificdreamer \cite{wang2024prolificdreamer} exhibit the best quality, authenticity, and correspondence respectively. The reasons for the subjective score differences among different models: From Figure 9 in the manuscript, it can be seen that the best quality, authenticity, and correspondence are Prolificdreamer, Magic3D, and Prolificdreamer respectively. Prolificdreamer uses variational score distillation to instead of score distillation sampling which used in other methods and solve the problems of over-saturation, over-smoothing, and low-diversity. So the Prolificdreamer has better quality and correspondence. Magic3D uses coarse-to-fine strategy to generate 3D asset and a sparse 3D hash grid structure to represent 3D asset, which can reduce the generation of floaters, making generated 3D asset more authenticity. 2) When the prompt is short (1 \& 2), the model is easy to generate high quality, authenticity, and correspondence 3D assets, However, as prompt length increases (3, 4 \& 5), text-to-3D generation models may struggle to meet the requirements of human preferences and the entire prompt, resulting in a decline in the quality, authenticity, and correspondence scores. Finally, when the prompt length is extreme long, the explicit descriptions make the quality, authenticity scores higher, while the correspondence scores are still lower than the prompt length of 1 \& 2. The reasons for subjective score differences in different prompt lengths: When the prompt length is short, the generated 3D asset is less constrained by the text, making it easy to achieve better text asset correspondence. However, as the length increases, the text-asset correspondence decreases; When the prompts are too long, a more detailed description can help the models generate better textures and geometry, resulting in better authenticity and quality.

\section{Proposed Method}

\begin{figure*}[t]
    \centering
    \includegraphics[width = 0.94\textwidth]{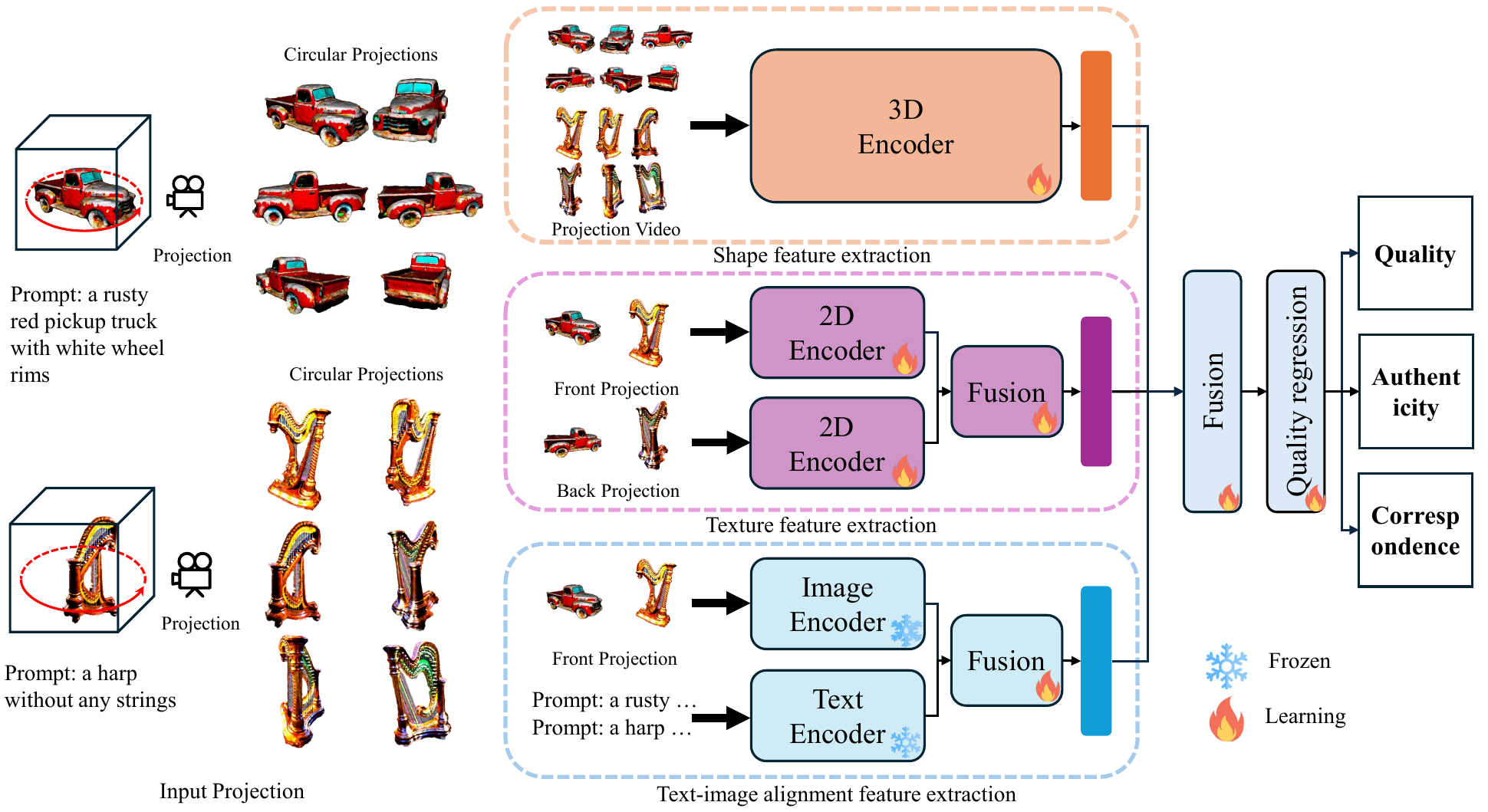}
    \caption{Illustration of our proposed T23DAQA method, which is divided into two stages. In the first stage, circular projection views are captured from the text-to-3D generated assets and then concatenated to form a video sequence. In the second stage, three distinct modules are employed to extract shape features, texture features, and text-image alignment features, respectively. Then these features are fused together to regress into quality, authenticity, and text-asset correspondence scores for comprehensive evaluation.}
    \label{fig:framework}
    \vspace{-0.5cm}
\end{figure*}

In this section, we introduce the architecture of the proposed text-to-3D asset quality assessment (T23DAQA) model in detail, as shown in Fig. \ref{fig:framework}. It is divided into two stages. In the first stage, we capture circular projections for the text-to-3D assets, and concatenate the projection views into videos. In the second stage, we first use the shape feature extraction module, texture feature extraction module, and text-image correspondence feature extraction module to extract the features related to human preferences respectively, and then fuse these features to regress into quality, authenticity, and text-asset correspondence scores for evaluation.
\vspace{-15pt}
\subsection{Projection Process}
                            Our T23DAQA model first represent the 3D asset into videos for the subsequent evaluation.
                            The reasons of choosing projection videos as the format to predict human preferences for 3D assets are given as follows. 1) text-to-3D asset generation methods usually adopt neural radiation field to represent the 3D asset, which is indirectly stored in MLP or voxel. This has resulted in a lack of a unified format for the generated 3D asset, making it difficult to be evaluated by 3D quality assessment methods. 2) The projection-based 3D quality assessment methods can be adapted to all kinds of 3D models, not only for the generated 3D asset but also for point cloud, mesh, voxel, \textit{etc}, since they infer the visual quality via the rendered projections. As shown in Fig. \ref{fig:framework}, we move the camera around the the generated 3D asset, then obtain a projection sequence and select $\mathbf{K}$ frames from it for subsequent processing, Given a text-to-3D asset $\mathbf{O}$, the projection process can be described as:

\begin{equation}
\begin{aligned}
     &\mathbf{P}  = R(\mathbf {O}), \\
    \mathbf{P}  = \{&\mathcal{P}_{k}|k =1,\cdots, \mathbf{K}\},
\end{aligned}
\end{equation}
\textcolor{black}{where $\mathbf{P}$ represents the set of select projection frames and $R(\cdot)$ stands for the rendering process, which determines the color of each pixel by calculating the density and color integral of the intersection of the ray passing through each pixel with the asset.}
\vspace{-13pt}
\subsection{Shape Feature Encoder}
The shape feature encoder is aimed to extract the 3D shape features of text-to-3d asset from the projection videos. Due to the use of implicit neural radiation fields to represent 3D asset, the shape of the T23DA is usually relatively smooth, and some may have floaters, which greatly affects the quality and authenticity of the T23DA. Therefore, we use a Swin3D-s\cite{liu2021swin} as the 3D shape encoder to extract the 3D shape feature from the projection video. This process can be represented as:
\begin{equation}
     f_{s}  =  E_{s}(\mathbf{P}),
\end{equation}
where $E_{s}$ and $f_{s}$ represents the projection video encoder and the obtained 3D shape features respectively.
\vspace{-13pt}
\subsection{Texture Feature Encoder}

The texture feature encoder is aimed to extract the texture feature of the text-to-3d asset from the image dimension, which represents the material and physical properties of the text-to-3d asset. If the texture feature is incompatible with the shape of 3D asset, the quality and authenticity of the T23DA are low. In order to extract the overall texture feature, we utilize Swin Transformer-small (Swin-s)\cite{liu2021swin} as the front projection image encoder and the back projection image encoder to extract the texture features. This process can be formulated as:
\begin{equation}
     f_{t}  = F_{t}( E_{t}^{f}(\mathcal{P}_{1}) , E_{t}^{b}(\mathcal{P}_{1+\frac{N}{2}}) ),
\end{equation}
\textcolor{black}{where $E_{t}^{f}$ and $E_{t}^{b}$ denote the encoders for the front and back projection images, respectively. $\mathcal{P}_{1}$ and $\mathcal{P}_{1+\frac{N}{2}}$ represent the front and back projection images. $F_{t}$ corresponds to the texture feature fusion module, while $f_{t}$ signifies the extracted texture features.}
\vspace{-15pt}
\subsection{Text-image Alignment Encoder}
The text-image alignment encoder is used to extract the text-image alignment feature. Following previous works, we use the pre-trained CLIP \cite{radford2021learning} image encoder $E_{ci}$ as the projection video frame encoder and the text encoder $E_{ct}$ as the prompt encoder. The two features extracted by these two encoders are fused to the alignment feature by alignment fusion module $F_{c}$. This process can be expressed as:

\begin{equation}
\begin{aligned}
     f_{c}^{i}  = E_{c}^{i} (\mathcal{P}_{1}) &, f_{c}^{t}  = E_{c}^{t}(W), \\
     f_{c}  = F_{c}& ( f_{c}^{i} , f_{c}^{t} ),
\end{aligned}
\end{equation}
\textcolor{black}{where $\mathcal{P}_{1}$ denotes the front projection image, and $W$ represents the prompt. The features $f_{c}^{i}$, $f_{c}^{t}$, and $f_{c}$ correspond to the image feature, prompt feature, and text-image alignment feature, respectively. During the training phase, the weights of the projection video frame encoder $E_{c}^{i}$ and the prompt  encoder $E_{c}^{t}$ are frozen, while only the alignment fusion module $F_{c}$ is trained.}

\begin{table*}[t]

\caption{Performance results of traditional handcrafted perceptual quality metrics and alignment metrics on our AIGC-T23DAQA database. [Key: {\bf\textcolor{red}{Best}, \bf\textcolor{blue}{Second Best}}]}
\label{tab:zero-shot}
\centering
\scalebox{1.0}{
\begin{tabular}{ll|rrr|rrr|rrr}
\toprule
\multicolumn{2}{c|}{\textbf{Dimension}}  & \multicolumn{3}{c|}{Authenticity}  & \multicolumn{3}{c|}{Correspondence} & \multicolumn{3}{c}{Quality}                                                 \\ \hline
\multicolumn{1}{c|}{\textbf{Type}}              & \multicolumn{1}{c|}{\textbf{Metric}} & \multicolumn{1}{c}{SRCC} & \multicolumn{1}{c}{KRCC} & \multicolumn{1}{c|}{PLCC} & \multicolumn{1}{c}{SRCC} & \multicolumn{1}{c}{KRCC} & \multicolumn{1}{c|}{PLCC} & \multicolumn{1}{c}{SRCC} & \multicolumn{1}{c}{KRCC} & \multicolumn{1}{c}{PLCC} \\ \hline
\multicolumn{1}{l|}{\multirow{10}{*}{\begin{tabular}[c]{@{}l@{}} NR-IQA \end{tabular}}}  & NIQE\cite{mittal2012making} & 0.1534 & 0.1270 & 0.1708 & 0.1272 & 0.0881 & 0.1755 & 0.0256 & 0.0209 & 0.1408 \\
\multicolumn{1}{l|}{} & ILNIQE\cite{zhang2015feature} & 0.0670 & 0.0556 & 0.0545 & 0.2764 & 0.1939 & 0.3092 & 0.2385 & 0.1727 & 0.2838 \\
\multicolumn{1}{l|}{} & BRISQUE\cite{mittal2012no} & 0.1224 & 0.0831 & 0.1461 & 0.1244 & 0.0852 & 0.1422 & 0.0884 & 0.0605 & 0.1100 \\
\multicolumn{1}{l|}{} & QAC\cite{xue2013learning} & 0.2472 & 0.1671 & 0.2662 & 0.1114 & 0.0921 & 0.0938 & 0.2198 & 0.1496 & 0.2415 \\
\multicolumn{1}{l|}{} & FISBLIM\cite{gu2013fisblim} & 0.1507 & 0.1049 & 0.1902 & 0.1297 & 0.0931 & 0.2015 & 0.3816 & 0.2764 & 0.4049 \\
\multicolumn{1}{l|}{} & BMPRI\cite{min2018blind} & 0.0680 & 0.0562 & 0.0740 & 0.0408 & 0.0282 & 0.1247 & 0.0486 & 0.0340 & 0.1063 \\
\multicolumn{1}{l|}{} & BPRI\cite{min2017blind} & 0.1322 & 0.0907 & 0.1752 & 0.0064 & 0.0041 & 0.1553 & 0.1354 & 0.0928 & 0.1655 \\
\multicolumn{1}{l|}{} & BPRI-PSS\cite{min2017blind} & 0.1296 & 0.0875 & 0.2508 & 0.1182 & 0.0792 & 0.3154 & 0.1866 & 0.1241 & 0.3436 \\
\multicolumn{1}{l|}{} & BPRI-LSSs\cite{min2017blind} & 0.0333 & 0.0220 & 0.0702 & 0.1089 & 0.0750 & 0.1465 & 0.0425 & 0.0277 & 0.0968 \\
\multicolumn{1}{l|}{} & BPRI-LSSn\cite{min2017blind} & 0.1345 & 0.1114 & 0.1332 & 0.2624 & 0.1855 & 0.3490 & 0.3136 & 0.2209 & 0.3551 \\ \hline
\multicolumn{1}{l|}{\multirow{5}{*}{\begin{tabular}[c]{@{}l@{}}Alignment\end{tabular}}} & CLIPScore\cite{hessel2021clipscore} & 0.4812 & 0.3324 & 0.5107 & 0.6053 & 0.4280 & 0.6584 & 0.5765 & 0.4057 & 0.5806 \\
\multicolumn{1}{l|}{} & HPS\cite{wu2023human} & 0.4393 & 0.3002 & 0.4589 & 0.5638 & 0.3922 & 0.5977 & 0.5876 & 0.4170 & 0.5856 \\
\multicolumn{1}{l|}{} & ImageReward\cite{xu2024imagereward} & \textcolor{blue}{0.5119} & \textcolor{blue}{0.3588}& \textcolor{blue}{0.5161} & \textcolor{red}{0.6604} & \textcolor{red}{0.4887} & \textcolor{red}{0.7027} & \textcolor{blue}{0.6585} & \textcolor{blue}{0.4752} & \textcolor{blue}{0.6469} \\
\multicolumn{1}{l|}{} & PickScore\cite{kirstain2024pick} & 0.4782 & 0.3335 & 0.5054 & 0.5396 & 0.3812 & 0.5792 & 0.5796 & 0.4115 & 0.5902 \\
\multicolumn{1}{l|}{} & ViCLIP\cite{wang2023internvid} & 0.4815 & 0.3327 & 0.5122 & \textcolor{blue}{0.6529} & \textcolor{blue}{0.4670}& \textcolor{blue}{0.6919} & 0.6235 & 0.4449 & 0.6304 \\ \hline
\multicolumn{1}{l|}{\multirow{3}{*}{\begin{tabular}[c]{@{}l@{}}\textcolor{black}{LMMQA}\end{tabular}}} & \textcolor{black}{Q-align} \cite{wu2023q} & 0.2339 & 0.1605 & 0.2941 & 0.1441 & 0.0997 & 0.2002 & 0.3906 & 0.2724 & 0.4302 \\
\multicolumn{1}{l|}{} & \textcolor{black}{T2I-Scorer} \cite{wu2024t2i} & \textcolor{red}{0.5449} & \textcolor{red}{0.3834} & \textcolor{red}{0.5567} & 0.4908 & 0.3411 & 0.5022 & \textcolor{red}{0.6771} & \textcolor{red}{0.4957} & \textcolor{red}{0.6835} \\
\multicolumn{1}{l|}{} & \textcolor{black}{VQAScore} \cite{lin2025evaluating} & 0.3701 & 0.2548 & 0.3849 & 0.5451 & 0.3805 & 0.5381 & 0.4373 & 0.3015 & 0.4390 \\

\bottomrule
\end{tabular}
}
\vspace{-0.5cm}
\end{table*}
\vspace{-13pt}
\subsection{Feature Fusion and Quality Regression}
The previous three modules extract the 3d shape, texture, and text-image alignment feature of the text-to-3D asset respectively. Finally, we concatenate these features to obtain the perception quality features $f$ for the text-to-3D asset:
\begin{equation}
     f  = \text{concatenate}(f_{c},f_{t},f_{f}),
\end{equation}

After extracting perception quality features through the designated feature extraction modules, we then map these features to preference scores using a regression module. In this model, we utilize a MLP as the regression module, due to its simplicity and effectiveness in terms of model complexity. The MLP architecture consists of three fully connected layers, with 1024 neurons in the first layer, 128 neurons in the second layer, and 3 neurons in the output layer. Consequently, through this process, we are able to derive quality, authenticity, and text-asset correspondence scores as follows:

\begin{equation}
     [\hat{Q}_{q},\hat{Q}_{a},\hat{Q}_{c}]  = F_{f}(f),
\end{equation}
where $F_{f}$ denotes the function of the three FC layers. $\hat{Q}_{q}$, $\hat{Q}_{a}$and $\hat{Q}_{c}$ are the predicted quality, authenticity, and text-image correspondence scores respectively.
\vspace{-13pt}
\subsection{Loss function}


In accordance with \cite{wu2022fastquality, li2020norm}, we employ linearity loss and monotonicity loss functions. The linearity loss function is used to force the predicted quality scores close to the quality labels, which can be regarded as Mean Squared Error loss with z-score normalization. We need to normalize the predicted scores vectors $\hat{Q}$ and ground truth label vectors $Q$ to obtain $\hat{S}$ and $S$ respectively.
The linearity loss can be described as:
\begin{equation}
\begin{aligned}
    L_{\text{lin}} =& \frac{((\hat{S} - S)^{2} + (\sum (\hat{S} * S)*\hat{S} - S)^{2})}{2}, \\
\end{aligned}
\end{equation}

while rank loss aids in enhancing the model's ability to discern the relative quality of projection videos, which can be formalized as:
\begin{equation}
L_{\text{rank}} = \sum \max((\hat{Q} - Q)sgn(\hat{Q} - Q), 0),
\end{equation}
where $sgn(\cdot)$ denotes the sign function. The composite loss function is formulated as follows:
\begin{equation}
L = \sum_{i} L_{\text{lin}} + \lambda \cdot L_{\text{rank}} \quad i \in \{ q, a, c\}, 
\end{equation}
Here, $\lambda$ denotes a hyper-parameter for balancing, which is set to 0.3 during the training phase. $q, a, c$ represent quality, authenticity, and text-asset correspondence respectively.

\section{Experimental Validation}
This section begins with a detailed outline of the experimental protocol, followed by an assessment of the performance of both conventional perception methods and the proposed approach on the AIGC-T23DAQA database. These perception models include traditional NR-IQA, NR-VQA, NR-MQA, NR-PCQA, LMMQA, T2IQA, T2VQA and alignment methods. Subsequently, we undertake ablation studies to illustrate the robustness and effectiveness of the proposed methodology.

\begin{table*}[tbph]

\caption{Performance results of learning-based metrics on our AIGC-T23DAQA database. [Key: {\bf\textcolor{red}{Best}, \bf\textcolor{blue}{Second Best}}]}
\label{tab:training}
\centering
\scalebox{1.0}{
\begin{tabular}{c|ll|rrr|rrr|rrr}
\toprule
\multirow{2}{*}{\begin{tabular}[c]{@{}l@{}} Index \end{tabular}} & \multicolumn{2}{c|}{\textbf{Dimension}}  & \multicolumn{3}{c|}{Authenticity}  & \multicolumn{3}{c|}{Correspondence} & \multicolumn{3}{c}{Quality}                                                 \\ \cline{2-12}
 & \multicolumn{1}{c|}{\textbf{Type}}              & \multicolumn{1}{c|}{\textbf{Metric}} & \multicolumn{1}{c}{SRCC} & \multicolumn{1}{c}{KRCC} & \multicolumn{1}{c|}{PLCC} & \multicolumn{1}{c}{SRCC} & \multicolumn{1}{c}{KRCC} & \multicolumn{1}{c|}{PLCC} & \multicolumn{1}{c}{SRCC} & \multicolumn{1}{c}{KRCC} & \multicolumn{1}{c}{PLCC} \\ \hline
A & \multicolumn{1}{l|}{\multirow{10}{*}{\begin{tabular}[c]{@{}l@{}} NR-IQA \end{tabular}}}  & Resnet-18\cite{he2016deep} & 0.5114 & 0.3618 & 0.5267 & 0.5652 & 0.4027 & 0.6240 & 0.6970 & 0.5166 & 0.7004 \\
B & \multicolumn{1}{l|}{} & Resnet-34\cite{he2016deep} & 0.5688 & 0.4047 & 0.5846 & 0.5794 & 0.4181 & 0.6325 & 0.7122 & 0.5288 & 0.7104 \\
C & \multicolumn{1}{l|}{} & Resnet-50\cite{he2016deep} & 0.4657 & 0.3354 & 0.4961 & 0.4750 & 0.3364 & 0.5629 & 0.6441 & 0.4772 & 0.6624 \\
D & \multicolumn{1}{l|}{} & Swin-T\cite{liu2021swin} & 0.5934 & 0.4273 & 0.6241 & 0.6360 & 0.4669 & 0.6951 & 0.7515 & 0.5734 & 0.7678 \\
E & \multicolumn{1}{l|}{} & Swin-S\cite{liu2021swin} &  0.6263 & 0.4541 & 0.6478 & 0.6434 & 0.4817 & 0.6983 & \textcolor{blue}{0.7652} & 0.5869 & \textcolor{blue}{0.7820} \\
F & \multicolumn{1}{l|}{} & Swin-B\cite{liu2021swin} & 0.6197 & 0.4483 & 0.6415 & 0.6431 & 0.4776 & 0.6995 & 0.7617 & 0.5844 & 0.7774 \\
G & \multicolumn{1}{l|}{} & Swin-L\cite{liu2021swin} & 0.6069 & 0.4396 & 0.6323 & 0.6539 & 0.4850 & 0.7132 & 0.7592 & 0.5810 & 0.7714 \\ 
H & \multicolumn{1}{l|}{} & CNNIQA\cite{kang2014convolutional} & 0.4281 & 0.2969 & 0.4332 & 0.5562 & 0.3932 & 0.6104 & 0.6776 & 0.4954 & 0.6658 \\
I & \multicolumn{1}{l|}{} & StairIQA\cite{sun2023blind} & 0.5002 & 0.3579 & 0.5375 & 0.4635 & 0.3360 & 0.5773 & 0.6373 & 0.4804 & 0.6715 \\ 
J & \multicolumn{1}{l|}{} & HyperIQA\cite{Su_2020_CVPR} & 0.6069 & 0.4396 & 0.6323 & 0.6539 & 0.4850 & 0.7132 & 0.7592 & 0.5810 & 0.7714 \\ 
\hline
K & \multicolumn{1}{l|}{\multirow{9}{*}{\begin{tabular}[c]{@{}l@{}}NR-VQA\end{tabular}}} & MC3-18\cite{tran2018closer} & 0.5702 & 0.4090 & 0.5948 & 0.6203 & 0.4554 & 0.6623 & 0.7421 & 0.5631 & 0.7528 \\
L & \multicolumn{1}{l|}{} & R2P1D-18\cite{tran2018closer} & 0.5864 & 0.4168 & 0.5903 & 0.6134 & 0.4520 & 0.6726 & 0.7423 & 0.5613 & 0.7474 \\
M & \multicolumn{1}{l|}{} & R3D-18\cite{tran2018closer} & 0.5869 & 0.4141 & 0.5951 & 0.5962 & 0.4327 & 0.6626 & 0.7430 & 0.5608 & 0.7466 \\
N & \multicolumn{1}{l|}{} & Swin3D-T\cite{liu2022video} & 0.6190 & 0.4521 & 0.6433 & 0.6517 & 0.4885 & 0.7034 & 0.7556 & 0.5842 & 0.7752 \\
O & \multicolumn{1}{l|}{} & Swin3D-S\cite{liu2022video} & 0.6317 & 0.4641 & 0.6517 & 0.6394 & 0.4795 & 0.7030 & 0.7579 & 0.5846 & 0.7768 \\
P & \multicolumn{1}{l|}{} & Swin3D-B\cite{liu2022video} & 0.6181 & 0.4502 & 0.6447 & 0.6294 & 0.4707 & 0.6973 & 0.7544 & 0.5809 & 0.7757 \\
Q &\multicolumn{1}{l|}{} & SimpleVQA\cite{sun2022a} & 0.6072 & 0.4545 & 0.6404 & 0.6102 & 0.4627 & 0.6971 & 0.7539 & \textcolor{blue}{0.5872} & 0.7712\\
R &\multicolumn{1}{l|}{} & Fast-VQA\cite{wu2022fastquality} & 0.6457 & 0.4690 & 0.6501 & 0.6477 & 0.4816 & 0.7071 & 0.7621 & 0.5813 & 0.7747 \\
S &\multicolumn{1}{l|}{} & DOVER\cite{wu2023dover} & \textcolor{blue}{0.6534} & \textcolor{blue}{0.4745} & \textcolor{blue}{0.6627} & \textcolor{blue}{0.6791} & 0.4954 & 0.7059 & 0.7508 & 0.5805 & 0.7708 \\ \hline
T & \multicolumn{1}{l|}{\multirow{2}{*}{\begin{tabular}[c]{@{}l@{}}NR-MQA\end{tabular}}} & NR-SVR\cite{abouelaziz2016no} & 0.3479 & 0.2637 & 0.4769 & 0.3163 & 0.3473 & 0.4921 & 0.5134 & 0.3904 & 0.5375 \\
U &\multicolumn{1}{l|}{} & NR-GRNN\cite{abouelaziz2016curvature} & 0.5613 & 0.3875 & 0.5336 & 0.4065 & 0.3025 & 0.5581 & 0.6052 & 0.4703 & 0.6074 \\\hline
V & \multicolumn{1}{l|}{\multirow{3}{*}{\begin{tabular}[c]{@{}l@{}}NR-PCQA\end{tabular}}} & 3D-NSS\cite{zhang2022no} & 0.3075 & 0.2190 & 0.3062 & 0.3969 & 0.2763 & 0.3094 & 0.5919 & 0.3937 & 0.5112 \\
W &\multicolumn{1}{l|}{} & ResSCNN\cite{liu2023point} & 0.4901 & 0.2445 & 0.4194 & 0.5965 & 0.3024 & 0.6785 & 0.6098 & 0.4961 & 0.6741 \\
X &\multicolumn{1}{l|}{} & IT-PCQA\cite{yang2022no} & 0.5663 & 0.3442 & 0.5950 & 0.4124 & 0.3551 & 0.5849 & 0.6405 & 0.4978 & 0.6797 \\ \hline
Y &\multicolumn{1}{l|}{\multirow{3}{*}{\begin{tabular}[c]{@{}l@{}}\textcolor{black}{T2IQA}\end{tabular}}} & \textcolor{black}{MA-AGIQA} \cite{wang2024large} & 0.6307 & 0.4558 & 0.6369 & 0.5965 & 0.4303 & 0.6713 &  0.7603 & 0.5767 & 0.7434 \\
Z &\multicolumn{1}{l|}{} & \textcolor{black}{MoE-AGIQA} \cite{yang2024moe} & 0.6386 & 0.4592 & 0.6396 & 0.6673 & 0.4999 & 0.6857 & 0.7350 & 0.5685 & 0.7454 \\
AA &\multicolumn{1}{l|}{} & \textcolor{black}{CLIP-AGIQA} \cite{tang2025clip}  & 0.6373 & 0.4689 & 0.6531 & 0.6739 & \textcolor{blue}{0.5057} & \textcolor{blue}{0.7197} & 0.7428 & 0.5683 & 0.7548 \\
\hline
AB &\multicolumn{1}{l|}{\multirow{2}{*}{\begin{tabular}[c]{@{}l@{}}\textcolor{black}{T2VQA}\end{tabular}}} & \textcolor{black}{T2VQA} \cite{kou2024subjective} & 0.6317 & 0.4365 & 0.6289 & 0.6489 & 0.4644 & 0.6704 & 0.7319 & 0.5526 & 0.7378 \\
AC &\multicolumn{1}{l|}{} & \textcolor{black}{TriVQA} \cite{qu2024exploring} & 0.6357 & 0.4588 & 0.6364 & 0.6353 & 0.4505 & 0.6717 & 0.7291 & 0.5331 & 0.7228 \\
\hline
AD & \multicolumn{1}{l|}{} & Proposed & \textcolor{red}{0.6728} & \textcolor{red}{0.4909} & \textcolor{red}{0.6840} & \textcolor{red}{0.7000} & \textcolor{red}{0.5157} & \textcolor{red}{0.7297} & \textcolor{red}{0.7853} & \textcolor{red}{0.5987} & \textcolor{red}{0.7828} \\
\bottomrule 
\end{tabular}
}
\vspace{-0.5cm}
\end{table*}
\vspace{-15pt}
\subsection{Experiment Protocol}

1) Baseline Algorithms:
In our evaluation, we incorporate a selection of representative NR-IQA, NR-VQA, NR-MQA, NR-PCQA algorithms, LMMQA, T2IQA, T2VQA and alignment methods as benchmarks for comparative analysis. These baseline methods encompass:
\begin{itemize}
    \item General NR-IQA methods: We test 20 baseline IQA methods categorized into two groups, including: traditional NR-IQA models and deep neural network (DNN) based NR-IQA models. For traditional NR-IQA, the selection models comprises NIQE \cite{mittal2012making}, ILNIQE \cite{zhang2015feature}, BRISQUE \cite{mittal2012no}, QAC \cite{xue2013learning}, FISBLIM \cite{gu2013fisblim}, BMPRI \cite{min2018blind}, BMPRI \cite{min2017blind}, BPRI-PSS \cite{min2017blind}, BPRI-LSSs \cite{min2017blind}, and BPRI-LSSn \cite{min2017blind}. In the realm of DNN-based NR-IQA, we consider Resnet-18 \cite{he2016deep}, Resnet-34 \cite{he2016deep}, Resnet-50 \cite{he2016deep}, Swin-T \cite{liu2021swin}, Swin-S \cite{liu2021swin}, Swin-B \cite{liu2021swin}, Swin-L \cite{liu2021swin}, CNNIQA \cite{kang2014convolutional}, HyperIQA \cite{sun2023blind}, and StairIQA \cite{sun2023blind}. These metrics represent widely used NR-IQA methodologies applied in practical applications.
    
    \item General NR-VQA methods: We test 9 baseline VQA methods on the constructed database including MC3-18\cite{tran2018closer}, R2P1D-18 \cite{tran2018closer}, R3D-18 \cite{tran2018closer}, Swin3D-T \cite{liu2022video}, Swin3D-S \cite{liu2022video}, Swin3D-B \cite{liu2022video}, SimpleVQA \cite{sun2022a}, Fast-VQA \cite{wu2022fastquality}, and DOVER \cite{wu2023dover}. These metrics serve as prevalent NR-VQA measures utilized in practical scenarios such as video coding and enhancement.

    \item General NR 3D quality assessment methods: We test 5 baseline 3DQA methods including NR-SVR \cite{abouelaziz2016no}, NR-GRNN \cite{abouelaziz2016curvature}, 3D-NSS \cite{zhang2022no}, ResSCNN \cite{liu2023point}, and IT-PCQA \cite{yang2022no}.
    
    \item Alignment methods: We select 5 baseline alignment methods: CLIPScore\cite{hessel2021clipscore}, HPS \cite{wu2023human}, ImageReward \cite{xu2024imagereward}, PickScore \cite{kirstain2024pick}, and ViCLIP \cite{wang2023internvid}. The first four metrics facilitate image-to-text alignment, and ViCLIP is tailored for video-to-text alignment applications.

    \textcolor{black}{\item LMMQA, T2IQA and T2VQA methods: We selected Q-align\cite{wu2023q}, T2I-Scorer\cite{wu2024t2i}, and VQAScore\cite{wu2024t2i} as representatives of LMMQA methods. Meanwhile, T2IQA and T2VQA mthods selecte MA-AGIQA\cite{wang2024large}, MoE-AGIQA\cite{yang2024moe}, CLIP-AGIQA\cite{tang2025clip} and T2VQA\cite{kou2024subjective}, TriVQA\cite{qu2024exploring} respectively.   }
    
\end{itemize}

2) Experimental and settings: 
For traditional NR-IQA and alignment methods and LMMQA, our evaluation encompasses the entire AIGC-T23DAQA database. For each projection video, these metrics predict scores for individual frames and derive the final prediction results by averaging these scores. However, for ViCLIP, the entire video is directly utilized to predict the final score.
As for CNN-based NR-IQA and general NR-VQA methods, we undertake fine-tuning on our AIGC-T23DAQA database. For SimpleVQA, Fast-VQA, DOVER, T2IQA, and T2VQA, we evaluate their performance using the provided open-source implementation. For the remaining algorithms, each projected video is segmented into an average of 12 segments. During training, one frame is randomly sampled from each segment, resulting in a total of 12 frames used as input. During testing, the first frame from each segment is selected. For IQA algorithms, the average score of the selected frames is computed as the final result.
Following the settings used in previous works \cite{wu2022fastquality, wu2023dover}, we partition the AIGC-T23DAQA database into training and test sets at a ratio of 4:1. Additionally, we conduct 10 random splits of the dataset and average the results to ensure unbiased performance comparison. We use the Adam optimizer with the initial learning rate set as $1e^{-4}$ and set the batch size as 4. The training process is stopped after 50 epochs. The resolution of input frames is rescaled to $224 \times 224$. 
\textcolor{black}{The image and text encoders used for text-image alignment feature extraction are from CLIP \cite{radford2021learning}. The 3D encoder for shape feature extraction is Swin3D-S\cite{liu2022video}, initialized with weights pretrained on the Kinetics dataset \cite{kay2017kinetics}. The 2D encoders for texture feature extraction are Swin-S\cite{liu2021swin}, initialized with weights pretrained on the ImageNet-1K dataset \cite{deng2009imagenet}.}
For NR-MQA and NR-PCQA, We export the generated 3D assets to mesh models using the Marching Cubes algorithm and convert the exported mesh models to point clouds using MeshLab. In the same time, we test several NR-MQA and NR-PCQA metrics on our proposed database.

3) Evaluation Criteria:
To evaluate the predictive accuracy of quality metrics, we employ three widely recognized global indicators, including Spearman’s Rank-Order Correlation Coefficient (SRCC), Kendall’s Rank-Order Correlation Coefficient (KRCC), and PLCC for assessing prediction monotonicity. Recognizing the potential presence of nonlinear mapping characteristics between objective scores and subjective scores, we apply score alignment by mapping the predicted values using the five-parameter logistic function, following the standard practice recommended in prior research \cite{sheikh2006statistical}:
\begin{equation}
\hat{Y}=\beta_{1}\left(0.5-\frac{1}{1+e^{\beta_{2}\left(Y-\beta_{3}\right)}}\right)+\beta_{4} Y+\beta_{5},
\end{equation}
where $\{\beta_{i} \mid i=1,2, \ldots, 5\}$ represent the parameters for fitting, $Y$ and $\hat{Y}$ stand for predicted and fitted scores respectively.



\begin{table*}[tbph]

\caption{Ablation study on AIGC-T23DAQA database. [Key: {\bf\textcolor{red}{Best}, \bf\textcolor{blue}{Second Best}}] The a-c represents the use of only text-image alignment feature extraction, texture feature extraction, and shape feature extraction module respectively, d-f represents not using the shape feature extraction, texture feature extraction, and text-image alignment feature extraction module, g uses all modules.}
\label{tab:ablation}
\centering
\scalebox{1.0}{
\begin{tabular}{c|rrr|rrr|rrr}
\toprule
\textbf{Dimension}  & \multicolumn{3}{c|}{Authenticity}  & \multicolumn{3}{c|}{Correspondence} & \multicolumn{3}{c}{Quality} \\ \hline
\textbf{Model} & \multicolumn{1}{c}{SRCC} & \multicolumn{1}{c}{KRCC} & \multicolumn{1}{c|}{PLCC} & \multicolumn{1}{c}{SRCC} & \multicolumn{1}{c}{KRCC} & \multicolumn{1}{c|}{PLCC} & \multicolumn{1}{c}{SRCC} & \multicolumn{1}{c}{KRCC} & \multicolumn{1}{c}{PLCC} \\ \hline

a & 0.6414 & 0.4654 & 0.6572 & 0.6805 & 0.5005 & 0.7073 & 0.7538 & 0.5683 & 0.7492 \\
b & 0.5882 & 0.4288 & 0.6092 & 0.6114 & 0.4477 & 0.6699 & 0.7467 & 0.5672 & 0.7539 \\
c & 0.5762 & 0.4162 & 0.6099 & 0.6070 & 0.4522 & 0.6852 & 0.7281 & 0.5544 & 0.7469 \\
d & 0.6369 & 0.4608 & 0.6504 & 0.6884 & 0.5080 & 0.7137 & 0.7627 & 0.5747 & 0.7565 \\
e & \textcolor{blue}{0.6665} & \textcolor{blue}{0.4853} & \textcolor{blue}{0.6792} & \textcolor{blue}{0.6997} & \textcolor{blue}{0.5082} & \textcolor{blue}{0.7277} & \textcolor{blue}{0.7766} & \textcolor{blue}{0.5891} & \textcolor{blue}{0.7729} \\
f & 0.5406 & 0.3962 & 0.5533 & 0.5790 & 0.4352 & 0.6301 & 0.6979 & 0.5315 & 0.7041 \\
g & \textcolor{red}{0.6728} & \textcolor{red}{0.4909} & \textcolor{red}{0.6840} & \textcolor{red}{0.7000} & \textcolor{red}{0.5157} & \textcolor{red}{0.7297} & \textcolor{red}{0.7853} & \textcolor{red}{0.5987} & \textcolor{red}{0.7828} \\
\bottomrule
\end{tabular}
}
\vspace{-0.5cm}
\end{table*}

\begin{figure*}[ht]
    \centering

    \subfigure[]{\begin{minipage}[t]{0.32\linewidth}
                \centering
                \includegraphics[width = 0.98\linewidth]{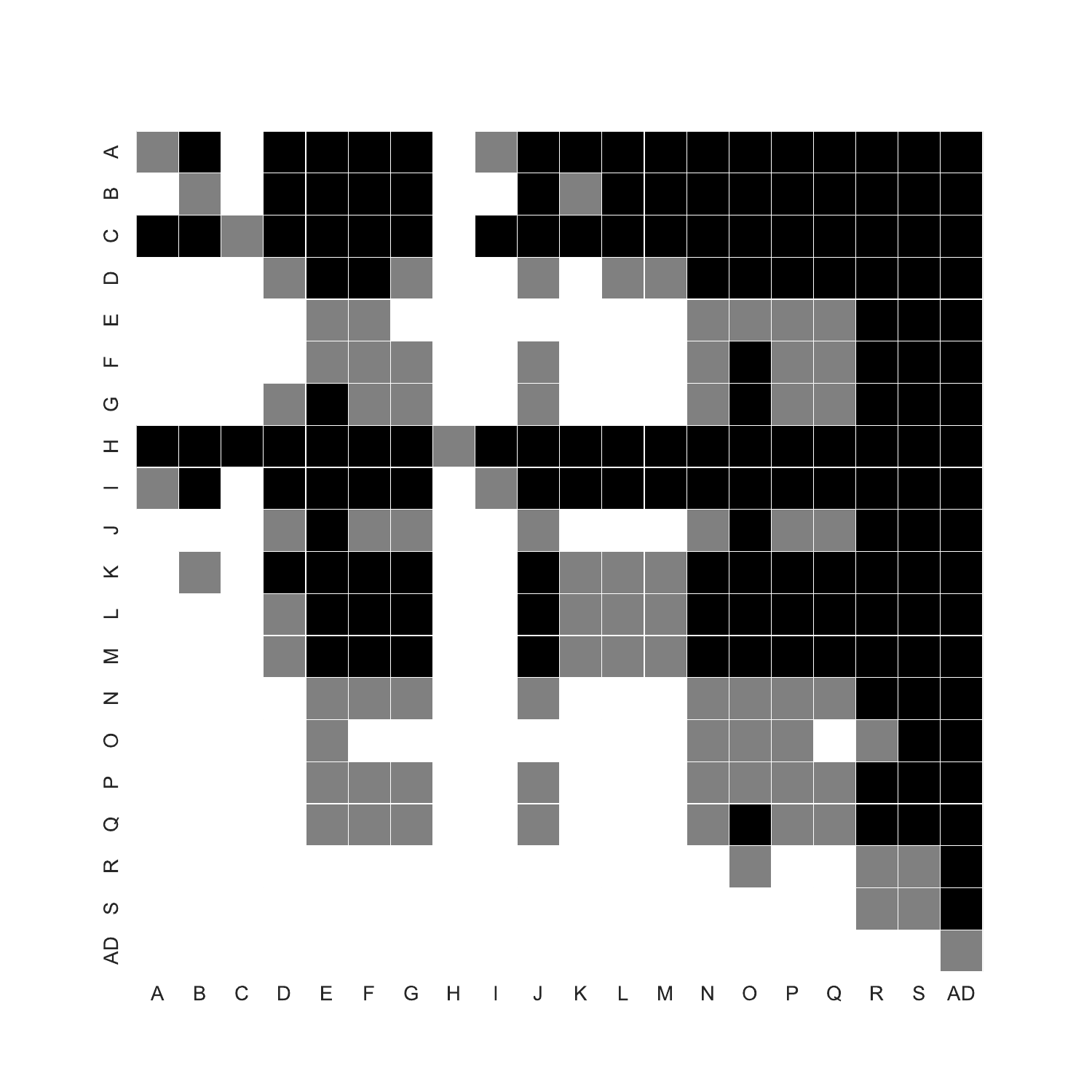}
                \end{minipage}}
    \subfigure[]{\begin{minipage}[t]{0.32\linewidth}
                \centering
                \includegraphics[width = 0.98\linewidth]{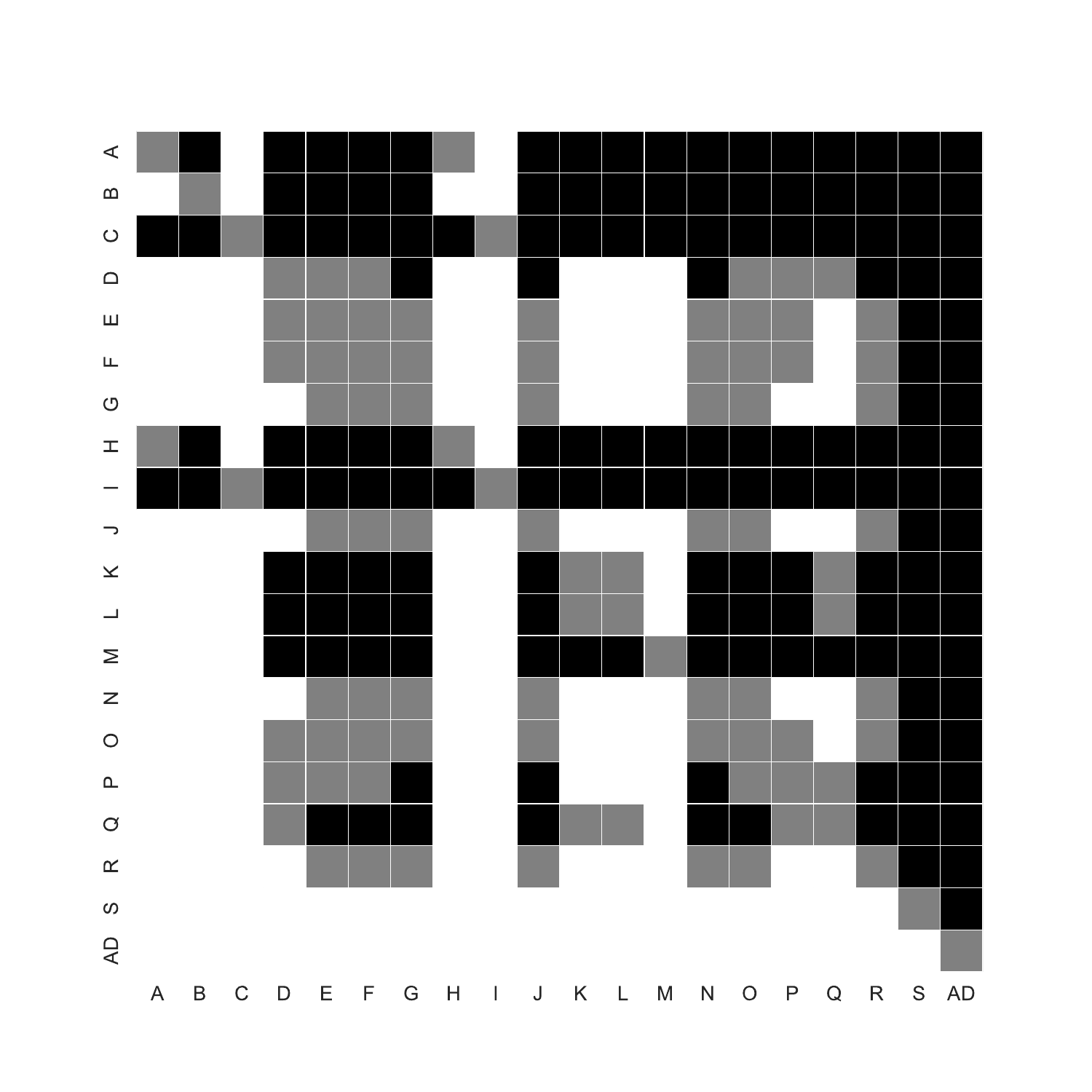}
                \end{minipage}}
    \subfigure[]{\begin{minipage}[t]{0.32\linewidth}
                \centering
                \includegraphics[width = 0.98\linewidth]{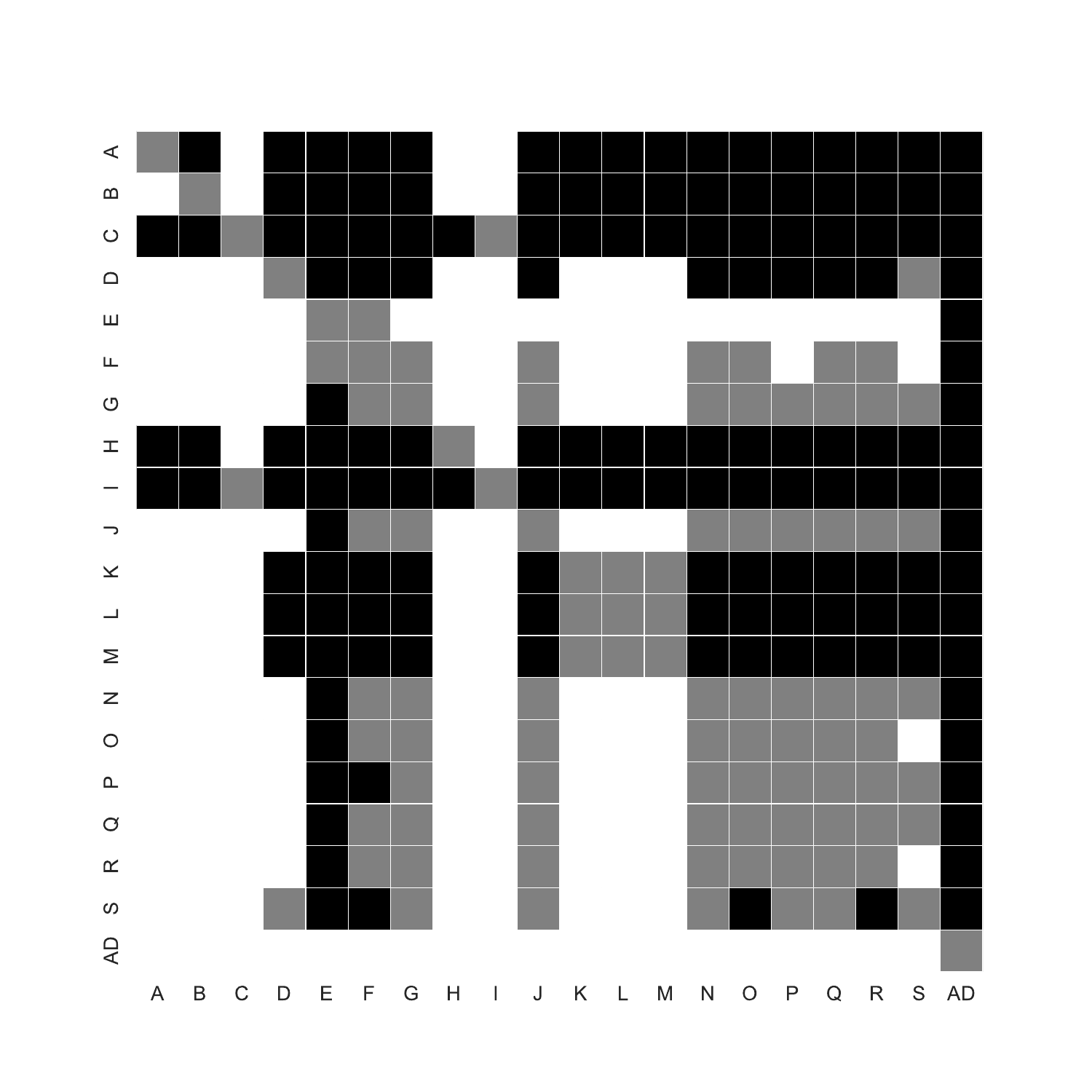}
                \end{minipage}}
                            
    \caption{The results of statistical tests on the AIGC-T23DAQA database. A black/white block indicates that the row method is inferior/superior to the column method, while a gray block signifies that there is no statistical difference between the row and column methods. The methods are identified by the same index as in Table \ref{tab:training}. }
    \label{fig:compare}
    \vspace{-0.5cm}
\end{figure*}
\vspace{-15pt}
\subsection{Experimental Results and Discussion}

Table \ref{tab:zero-shot} shows the performance results of various traditional NR-IQA methods and alignment methods on the established AIGC-T23DAQA database. From the results, we can get the following observation and conclusions:  1) Traditional NR-IQA methods exhibit relative poor performance. This is because NR-IQA methods predict image perception quality through handcrafted natural image texture features, which have a low correlation with the perception quality of generated 3D assets. 2) The quality of generated 3D asset is more correlated with traditional NR-IQA methods than authenticity and text 3D asset correspondence, which manifests that the authenticity and text-asset correspondence are two unique factors significantly different with the quality.  3) Alignment methods achieve commendable results due to the strong correlation between the generated 3D assets and prompts. Hence, employing a text-image alignment model can significantly enhance prediction accuracy. 4) Predicting the text-3d asset correspondence of generated 3D asset can assist in predicting its authenticity and quality. Table \ref{tab:training} showcases the performance results of different DNN-based NR-IQA methods, NR-VQA methods, NR-MQA methods, NR-PCQA methods and our proposed method on the proposed AIGC-T23DAQA database. The observations and conclusions are summarized as follows. 1) Our proposed method surpasses all baselines in terms of SRCC, KRCC, and PLCC, which demonstrates the effectiveness of our proposed method. 2) Overall, NR-VQA methods outperform NR-IQA methods, primarily due to their ability to extract 3D shape features from projection videos of generated 3D assets. Moreover, to gain further insight into the performance of the proposed method, we also conduct a significance-statistic test. \textcolor{black}{3) The performance of NR-MQA and NR-PCQA methods is worse than that of NR-VQA. The main reason for this is than the generated 3D asset utilizes implicit representations, such as occupancy fields or signed distance functions. Converting these to explicit representations (meshes or point clouds) will introduce distortions and loss of detail. This conversion process may adversely affect the quality assessment, as NR-MQA and NR-PCQA are sensitive to such distortions.} Our experiment setup follows the same procedure outlined in \cite{sheikh2006statistical} and evaluates the significance of the correlation between the predicted quality, authenticity, and correspondence scores and the subjective ratings. All possible pairs of models are tested and the results are displayed in Fig. \ref{fig:compare}. The results reveal that our method is significantly better than the other 10 NR-IQA methods and 9 NR-VQA methods.

\vspace{-13pt}
\subsection{Ablation Study}
To demonstrate the effectiveness of each module in our proposed method, we further conduct ablation experiments, and the results are presented in Table \ref{tab:ablation}. The ``a-c'' denote the utilization of only the text-image alignment feature extraction, texture feature extraction, and shape feature extraction modules respectively, while ``d-f'' represents the absence of shape feature extraction, texture feature extraction, and text-image alignment feature extraction modules respectively. The ``g'' configuration employs all modules. From the results, we draw the following conclusions. 1) All three proposed modules are effective for boosting the performance, while the text-image alignment feature extraction module playing the most significant role. This prominence can be attributed to the strong relationship between text-to-3d assets and their prompts, enabling the prompts to substantially contribute to predicting the perception quality of text-to-3d assets. \textcolor{black}{For the traditional 3D quality assessment, the focus is primarily on the aspects such as geometry and texture quality. While the T23DAQA need to not only assess the geometry and texture quality of generated 3D assets, but also a comprehensive evaluation of the alignment between text and 3D assets, encompassing semantic consistency, style matching. Therefore, the text-image alignment feature is the most important feature.} 2) The 3D shape feature extraction module and texture feature extraction module can effectively extract perceptually relevant features from the projected video and front and back projected images, respectively. Consequently, the two modules can enhance the accuracy of quality, authenticity, and correspondence prediction.
\vspace{-13pt}

\section{Conclusion}
AIGC is currently a hot research topic, and text-to-3D asset generation is an important part in this field. This paper contributes to the first study of text-to-3D asset quality assessment, which is a significant achievement to the area. Specifically, this paper addresses this problem by introducing the largest T23DAQA database to date, named AIGC-T23DAQA. Subsequently, a novel projection-based evaluator for better text-to-3D asset quality assessment, which leverages a 3D encoder, two 2D encoders, and multi-modality foundation models to extract 3D shape features, texture features, and text 3D asset correspondence features from projection videos and fuses to generate preference scores from the perspectives of quality, authenticity, and text-asset correspondence. Experimental results underscore the superiority of our proposed T23DAQA method, surpassing state-of-the-art NR-IQA, NR-VQA, NR-MQA, and NR-PCQA, LMMQA, T2IQA, T2VQA models. Ablation experiments further confirm the effectiveness of the proposed submodule.

\bibliographystyle{IEEEtran}
\bibliography{output}
 


\vfill

\end{document}